\documentclass{article}

\usepackage{microtype}
\usepackage{graphicx}
\usepackage{subfigure}
\usepackage{booktabs} % for professional tables

\usepackage{hyperref}

\usepackage[accepted]{icml2024}

\usepackage{amsmath}
\usepackage{amssymb}
\usepackage{mathtools}
\usepackage{amsthm}

\usepackage[capitalize,noabbrev]{cleveref}

\usepackage{colortbl}
\definecolor{lightblue}{rgb}{0.8,0.9,1}
\usepackage{nicematrix}
\usepackage{makecell}
\usepackage{multirow}
\usepackage[T1]{fontenc}
\usepackage{longtable}
\usepackage{titletoc}

\theoremstyle{plain}

\theoremstyle{definition}

\theoremstyle{remark}

\usepackage[textsize=tiny]{todonotes}

\icmltitlerunning{Crafting Large Language Models for Enhanced Interpretability}

\begin{document}

\twocolumn[
\icmltitle{Crafting Large Language Models for Enhanced Interpretability}




\begin{icmlauthorlist}
\icmlauthor{Chung-En Sun}{yyy}
\icmlauthor{Tuomas Oikarinen}{yyy}
\icmlauthor{Tsui-Wei Weng}{yyy}
\end{icmlauthorlist}

\icmlaffiliation{yyy}{UC San Diego}

\icmlcorrespondingauthor{Chung-En Sun}{cesun@ucsd.edu}
\icmlcorrespondingauthor{Tsui-Wei Weng}{lweng@ucsd.edu}

\icmlkeywords{Machine Learning, ICML}

\vskip 0.3in
]



\printAffiliationsAndNotice{}  % leave blank if no need to mention equal contribution

\begin{abstract}
We introduce the Concept Bottleneck Large Language Model (CB-LLM), a pioneering approach to creating inherently interpretable Large Language Models (LLMs). Unlike traditional black-box LLMs that rely on post-hoc interpretation methods with limited neuron function insights, CB-LLM sets a new standard with its built-in interpretability, scalability, and ability to provide clear, accurate explanations. This innovation not only advances transparency in language models but also enhances their effectiveness. Our unique Automatic Concept Correction (ACC) strategy successfully narrows the performance gap with conventional black-box LLMs, positioning CB-LLM as a model that combines the high accuracy of traditional LLMs with the added benefit of clear interpretability --- a feature markedly absent in existing LLMs.

\end{abstract}

\section{Introduction}
\label{sec:introduction}
Large Language Models (LLMs), such as BERT \cite{bert} and GPT3 \cite{gpt3}, have become instrumental in advancing  Natural Language Processing (NLP) tasks. However, the inherent opacity of these models poses significant challenges in ensuring their reliability, particularly when outcomes are based on unclear or flawed reasoning. This lack of transparency complicates the effort to debug and improve these models. 

% Despite the power of Large Language Models (LLMs) for solving many Natural Language Processing (NLP) tasks, the inner workings of LLMs remain opaque to human understanding. With the widespread use of LLMs (i.e. BERT \cite{bert} and GPT3 \cite{gpt3}), ensuring the reliability of these models has emerged as a significant concern. The last thing we want is for LLMs to produce outcomes rooted in incorrect or fathomless knowledge, posing challenges in debugging and detection.

Recent efforts in the field have primarily focused on post-hoc interpretations of neurons within LLMs \cite{openai, neuronsnlp, indivneurons}. Given a learned LLM, these studies aim to elucidate the inner workings of black-box language models by finding post-hoc explanations for neurons \cite{openai, leenlpexp, neuronsnlp, indivneurons}. Nevertheless, the explanations derived from these methods often do not accurately align with the activation behaviors of the neurons. Moreover, they often fall short in offering clear directions for model editing or debugging, thereby limiting their practical application in correcting outputs. 

% Recently, many studies have worked on interpreting the neurons in LLMs \cite{openai, neuronsnlp, indivneurons}. They focus on finding post hoc explanations for neurons in a given black-box language model. Besides, given these explanations, it is not clear how to perform model editing or debugging to correct the outputs, thereby limiting the practical utility of the interpretations.

% Nevertheless, the explanations derived from these methods often do not align with the activation behaviors of the neurons.

Motivated by these limitations, we propose the Concept Bottleneck Large Language Model (CB-LLM) -- the first concept bottleneck model (CBM) for NLP tasks. Our method can transform any pretrained language model into a CBM with an inherently interpretable concept bottleneck layer and a prediction layer. Our contributions are as follows:
\begin{itemize}
    \item We present the first CBM framework for LLMs that scales to large text classification benchmarks. Our CB-LLM encapsulates the best of both worlds: it matches the high accuracy of traditional black-box models across multiple datasets while also offering clear interpretability, a feature absent in existing LLMs.
        % \item Our CB-LLM achieves the same accuracy as black-box language models across multiple datasets.
    % \item Our CB-LLM is naturally interpretable and provides faithful interpretations, which closely align with human understanding and reasoning.
    \item Our proposed pipeline to build CB-LLM is fully automatic and efficient: it eliminates the need for human-annotated concept labels, and the computational cost is almost the same as the standard fine-tuning. Furthermore, our proposed Automatic Concept Correction (ACC) strategy efficiently boosts the performance of our CB-LLM in terms of both accuracy and faithfulness evaluation.
    \item Our CB-LLM matches the accuracy of the standard black-box models and achieves a $1.39\times$ higher average rating compared to the random baseline on the faithfulness evaluation. This suggests that our CB-LLM provides high-quality interpretability without sacrificing performance. 
    \end{itemize}

\section{Background and related works}
\label{sec:background}
\paragraph{Post-hoc neuron analysis for NLP.} Post-hoc analysis is the most popular method for comprehending the inner workings of black-box language models. Traditionally, this analytical approach comprises several methodological categories, each offering distinctive insights. Visualizing-based methods \cite{visualizing} involve the graphical representation of neuron activations and manually identify the underlying concepts. Corpus-based methods \cite{reprnn, indivneurons} involve aggregating statistical information derived from data activations to uncover the roles of neurons. Probing-based methods \cite{neuronsnlp} entail training classifiers over activations to pinpoint neurons associated with predefined linguistic concepts. Causation-based approaches \cite{ablation} identify neurons through controlling perturbations and observing prediction change.

Recently, with the advent of Large Language Models (LLMs) such as GPT, \cite{openai} proposes utilizing GPT4 to generate explanations for GPT2 neurons and simulating the real neuron activations. The subsequent comparison of simulated and actual activations facilitates an evaluation of the quality of explanations. Additionally, \cite{LLMposthoc} delves into the capability of LLMs to explain other predictive models. Given a dataset and model to explain, they perform in-context learning (ICL) to prompt LLMs to give explanations and highlight that LLMs can generate faithful explanations and consistently outperform previous post-hoc methods. 

While the notion of utilizing LLMs for post-hoc explanations appears promising, the challenge lies in the fact that the intricate nature of a neuron from a black-box language model may not be effectively articulated through natural language, potentially resulting in oversimplification and overlooking complex behaviors. Moreover, the considerable computational resources required for this approach restrict its applicability to explaining only a small fraction of neurons in a language model. In contrast, our proposed CB-LLM offers intrinsic interpretability without the need to obtain post-hoc interpretations.
 
\paragraph{CBM in image classification.} Recently, Concept Bottleneck Models (CBMs) \cite{cbm} have been revisited in the context of image classification tasks. CBMs incorporate a concept bottleneck layer (CBL), where individual neurons are designed to learn specific concepts that are interpretable by humans. CBL is then followed by the final fully connected layer responsible for making predictions. Training a CBM typically involves utilizing human-annotated concept labels, enabling the CBL to make multilabel predictions for these concepts when presented with an image. However, a significant limitation arises from the computational expense of constructing an entire CBM from scratch and the dependency on human-annotated concept labels. Addressing this challenge, \cite{phcbm} introduced a computationally economical algorithm that transforms any image classifier into a CBM. This transformation is achieved by leveraging Concept Activation Vectors (CAV) \cite{cav} or the multi-modal CLIP (Contrastive Language-Image Pretraining) model \cite{clip}. It's important to note that their approach requires either concept labels to obtain CAV or restricting the backbone to the CLIP image encoder if concept labels are unavailable, which does not fully resolve the limitation. Recognizing this constraint, \cite{lfcbm} proposed a Label-free CBM, which learns a CBM without relying on concept labels by leveraging the interpretability tool CLIP-Dissect \cite{clipdissect}.

Despite the extensive exploration of CBMs in the field of image classification tasks, to the best of our knowledge, there is still no CBM that scales to large NLP benchmarks. Consequently, our work focuses on learning an efficient, automated, and high-performance CBM specifically for LLMs.

\begin{figure*}[!t]
\centering
\includegraphics[width=0.95\textwidth]{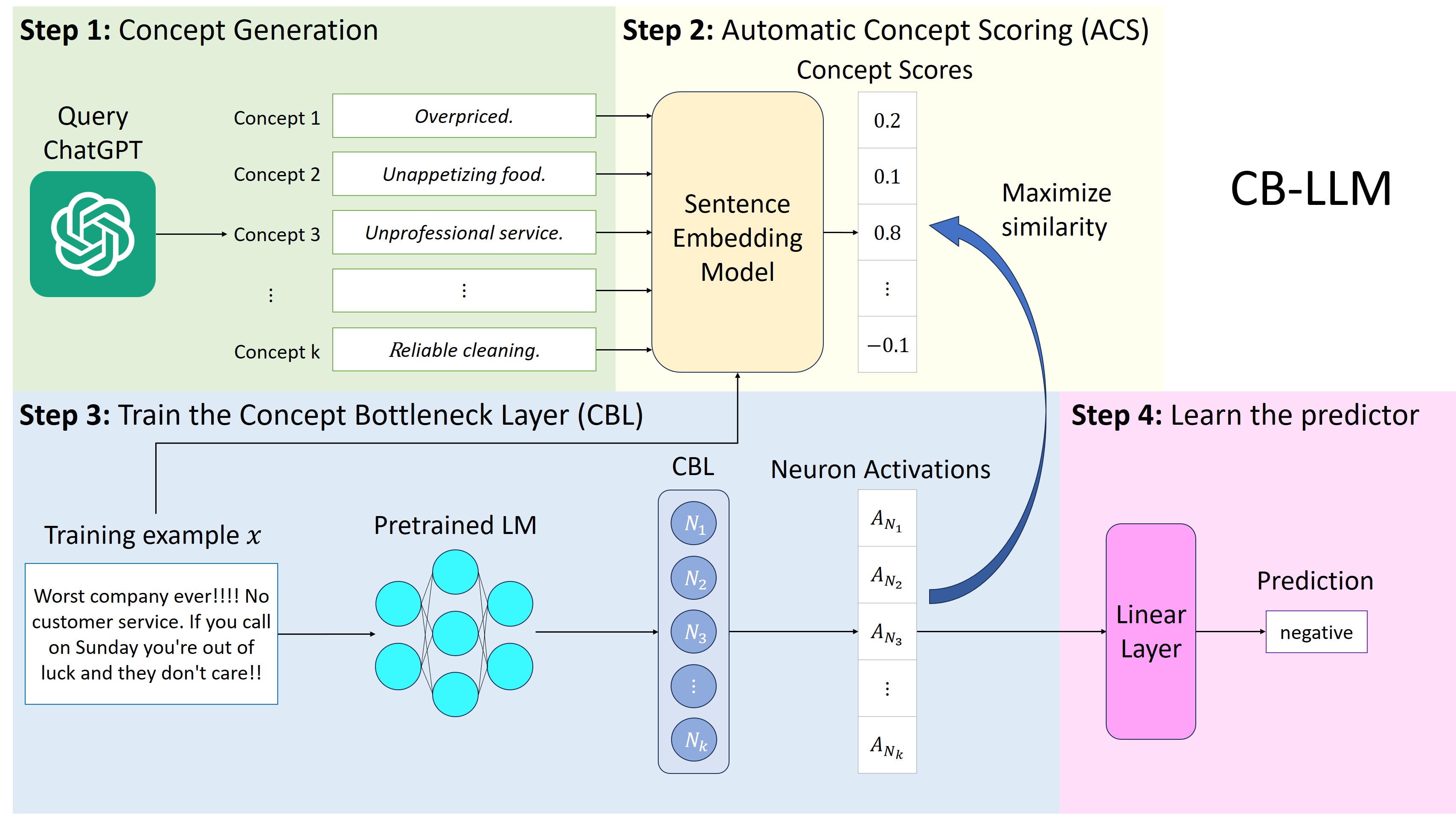}
\vspace{-10pt}
\caption{The overview of our CB-LLM.}
\label{fig:cbllm}
\end{figure*}

\paragraph{Sentence embedding models with contrastive learning.}
Contrastive learning has emerged as a predominant technique in training sentence embeddings, replacing the traditional approach of augmenting word2vec \cite{word2vec} with n-gram embeddings. 
%Given a set of positive paired sample $\{x_i,x_i^+\}^M_{i=1}$, where $x_i$ and $x_i^+$ are semantically related, contrastive learning pulls the embedding of the positive pairs closer by minimizing cross-entropy loss with in-batch negatives: $\mathcal{L}=-\sum^M_{i=1}\log{\dfrac{\exp^{\cos(h_i,h_i^+)/\tau}}{\sum^N_{j=1}\exp^{\cos(h_i,h_j^+)/\tau}}}$, where $(h_i,h_i^+)$ is the embedding of $(x_i,x_i^+)$, $N$ is the mini-batch size, $\cos$ is the cosine similarity and $\tau$ is the temperature. In unsupervised contrastive learning, $x_i^+$ is obtained by applying augmentation to $x_i$ such as word deletion, substitution, or reordering. 
A noteworthy method, SimCSE \cite{simcse}, has demonstrated success in semantic textual similarity (STS) tasks. They employ supervised contrastive learning to train the sentence embedding model with Natural Language Inference (NLI) datasets. This involves using entailment pairs as positive instances and contradiction pairs as hard negatives.
% Recently, \cite{angle} proposed AnglE, which suggests utilizing angle optimization in a complex space to replace the cosine similarity in contrastive learning. This approach mitigates the adverse effects of the saturation zone in cosine similarity, yielding state-of-the-art performance.

In our work, we leverage sentence embedding models trained with contrastive learning for Automatic Concept Scoring (ACS). This method yields high-quality concept scores without any human effort, which is a key step in building CB-LLM.

% \paragraph{Model editing for LLMs.} As the computational cost of training LLMs increases significantly, model editing has emerged as a novel technique to efficiently adjust the behavior of a base model on the desired examples, while avoiding impacting the model's behavior on other samples. \cite{knowledgeNeurons} propose an attribution method to identify knowledge neurons that express factual knowledge. \cite{memit} proposed MEMIT, which applied causal mediation analysis to locate the editing area and directly updated a language model with multiple
% memories.

% While recent studies demonstrate the possibility of editing LLMs, these methods require locating and optimizing neurons to incorporate new memories. In contrast, our CB-LLM offers a much simpler editing process to achieve desired outputs due to the interpretable nature of CBM.

\section{CB-LLMs: Building Interpretable Large Language Models}
\label{sec: concept bottleneck large language model}
% In the field of NLP, it is common to fine-tune a pretrained model for specific tasks. However, existing fine-tuned language models lack interpretability despite their good performance. Hence, in this section, we introduce how to transform a pretrained language model into a CB-LLM through our special fine-tuning strategy. 

Existing large language models (LLMs), despite their impressive performance, often lack interpretability. This section introduces a methodology that addresses this critical gap by employing a novel strategy. Our method transforms black-box pretrained models into interpretable entities, specifically converting them into Concept Bottleneck Large Language Models (CB-LLMs). This transformation significantly boosts interpretability without sacrificing performance. While our approach is adaptable to both fine-tuning pretrained models and training LLMs from scratch, we predominantly focus on building from pretrained models, as fine-tuning is a more common practice in NLP due to computational costs.

Our proposed method consists of four steps and is illustrated in Figure \ref{fig:cbllm}:
\begin{enumerate}
    \item \textbf{Concept Generation:} given a text classification task, generate a concept set for each class by prompting modern language models.
    \item \textbf{Automatic concept scoring (ACS):} leverage sentence embedding models to measure the similarity between each concept in the concept set and each text sample in the dataset.
    \item \textbf{Train the Concept Bottleneck Layer:} learn the concept mapping from uninterpretable features to human-interpretable concepts by maximizing the similarity between the neuron activations and the concept scores.
    \item \textbf{Learn the predictor:} train the final linear layer to make predictions for the downstream tasks.
\end{enumerate}
The details of steps 1 and 2 can be found in Section \ref{sec: concept generation} and \ref{sec: concept labeling} respectively. The details of steps 3 and 4 can be found in Section \ref{sec: learning CBLLM}.

\subsection{Concept generation}
\label{sec: concept generation}
The first step is to generate a set of concepts related to the downstream task. To automate this process, we leverage ChatGPT \cite{chatgpt} as a replacement for the domain experts. For any text classification dataset $\mathcal{D}$ with $n$ classes/labels, we prompt ChatGPT to generate the concept subset $\mathcal{S}_i$ for each class $i$. Then, the concept set $\mathcal{C}$ is the union of $\mathcal{S}_i$, $\mathcal{C}=\bigcup_{i=0}^{n-1} \mathcal{S}_i$. The following is the template we use to prompt ChatGPT to get $\mathcal{S}_i$:

\begin{itemize}
    \item "Here are some examples of key features that are often present in a \{\emph{class}\}. Each feature is shown between the tag \textless example\textgreater\textless /example\textgreater.
    \begin{itemize}
        \item \textless example\textgreater\{\emph{example 1}\}\textless /example\textgreater
        \item \textless example\textgreater\{\emph{example 2}\}\textless /example\textgreater
        \item \textless example\textgreater\{\emph{example 3}\}\textless /example\textgreater
        \item \textless example\textgreater\{\emph{example 4}\}\textless /example\textgreater
    \end{itemize}
    List \{\emph{concept size per class $|\mathcal{S}_i|$}\} other different important features that are often present in a \{\emph{class}\}. Need to follow the template above, i.e.\textless example\textgreater features\textless /example\textgreater."
\end{itemize}

We use four human-designed concepts as examples for in-context learning. This prompting style requires only $n$ queries to ChatGPT to obtain the full concept set and can be done efficiently through the web interface provided by OpenAI. More prompting details can be found in Appendix \ref{sec:detailed prompts}.

\subsection{Automatic Concept Scoring (ACS)}
\label{sec: concept labeling}

\begin{figure*}[!t]
\centering
\includegraphics[width=0.7\textwidth]{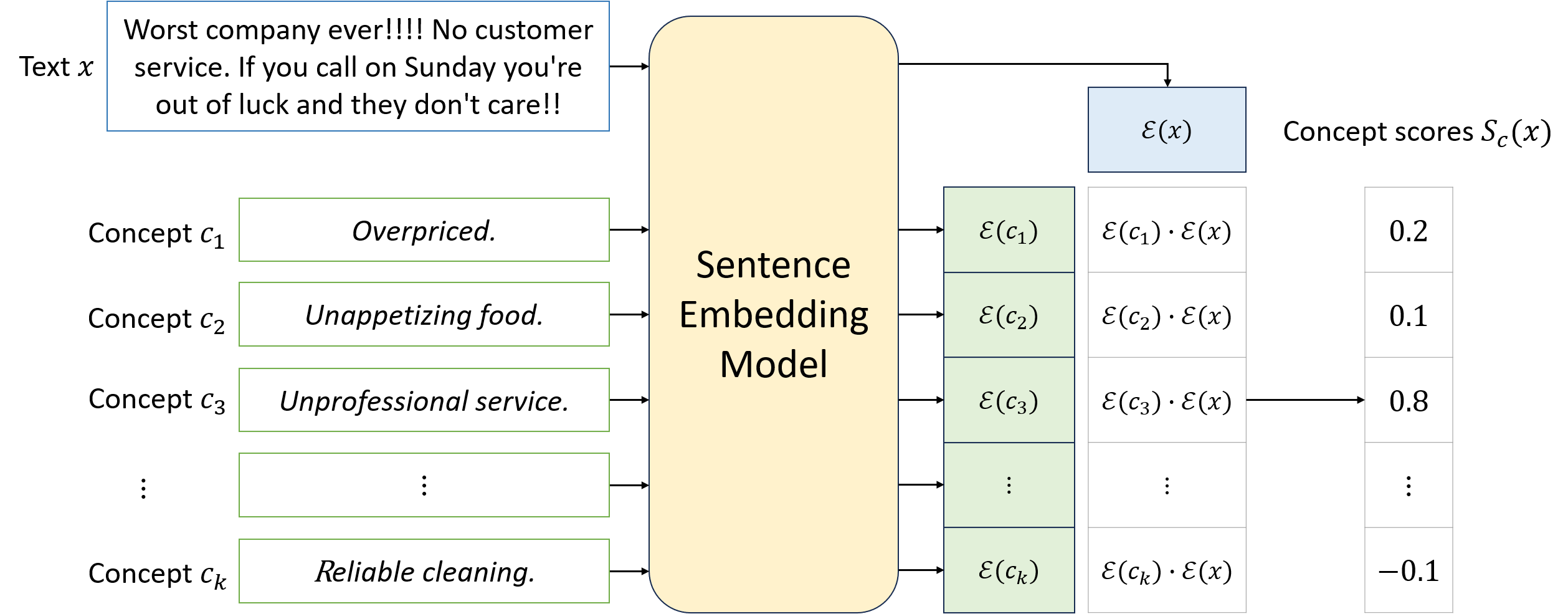}
\vspace{-10pt}
\caption{The process of Automatic Concept Scoring (ACS) through sentence embedding models.}
\label{fig:sentence embedding}
\end{figure*}

After generating the concept set $\mathcal{C}$, the next step is to obtain the concept labels for a given text sample $x$ in dataset $\mathcal{D}$. Typically, this stage requires involving domain experts and can be time-consuming. To overcome this challenge, we propose an automatic scoring strategy by utilizing sentence embedding models, which can measure the similarity between each concept and any text sample $x$. We name this strategy as Automatic Concept Scoring (ACS) and describe the details below.

For any sentence embedding model $\mathcal{E}$ that encodes a text sample into a fixed-size embedding, we calculate the concept scores $S_c(x)\in\mathbb{R}^k$ for text sample $x$ by calculating the following:
\begin{equation}
\label{e:sc}
    S_c(x) = [\mathcal{E}(c_1)\cdot\mathcal{E}(x),\mathcal{E}(c_2)\cdot\mathcal{E}(x),...,\mathcal{E}(c_k)\cdot\mathcal{E}(x)]^\top,
\end{equation}
where $\mathcal{E}(x)\in\mathbb{R}^d$ denotes the text embedding generated by $\mathcal{E}$, $c_i$ is the $i$-th concept in the concept set $\mathcal{C}$, and $k$ is the size of the concept set. Each component of the vector $S_c(x)$ represents the degree of association between the text $x$ and the concept $c_i$. This vector will be used as the learning target for CBL in the next section. The process of getting $S_c(x)$ is shown in Figure \ref{fig:sentence embedding}.

It's worth noting that, for a dataset with $m$ text examples $\mathcal{D}=\{x_1,...,x_m\}$ and a concept set with $k$ concepts $\mathcal{C}=\{c_1,...,c_k\}$, our ACS strategy requires only $m+k$ inferences to label the entire dataset. This stands in stark contrast to the more expensive alternative of utilizing zero-shot classification models trained with NLI datasets, which would require $mk$ inferences to label each pair of $(x_i,c_j), i\in\{1,...,m\}, j\in\{1,...,k\}$.

We use the off-the-shelf sentence embedding models \textbf{\texttt{all-mpnet-base-v2}} from Huggingface \cite{huggingface} for ACS. \textbf{\texttt{all-mpnet-base-v2}} is fine-tuned from pretrained \textbf{\texttt{MPNet}} model \cite{mpnet} with self-supervised contrastive learning objective using 1 billion sentence pairs. It serves as a computationally efficient option for ACS.

    % \item \textbf{\texttt{angle-llama-7b-nli-v2} (angle):} \cite{angle} suggested using angle optimization in a complex space for supervised contrastive learning and proposed the state-of-the-art sentence embedding model in the domain of Semantic Textual Similarity (STS). This model is finetuned from \texttt{Llama-7b} model \cite{llama} using multiple NLI datasets. While coming with a higher computational cost, it excels in capturing semantic meaning.

\subsection{Learning CB-LLM}
\label{sec: learning CBLLM}
After ACS, we have the concept scores $S_c(x)$ for every text example $x$ in dataset $\mathcal{D}$. Our CB-LLM is trained based on these concept scores and the class labels of $\mathcal{D}$. The training process unfolds in two sequential steps: first, a Concept Bottleneck Layer (CBL) is trained to learn the concepts, and subsequently, a linear predictor is trained to make the final predictions.

\paragraph{Training the concept bottleneck layer (CBL):} In this step, the goal is to force the neurons in CBL to activate in correlation with the pattern of concept scores. We first send the text sample $x$ into a pretrained LM $f_\textrm{LM}$ and use CLS pooling to get a fix size embedding $f_\textrm{LM}(x)\in\mathbb{R}^d$. Then, the CBL $f_\textrm{CBL}$ projects the embeddings into a $k$ dimensional interpretable embedding $f_\textrm{CBL}(f_\textrm{LM}(x))\in\mathbb{R}^k$. Note that $f_\textrm{CBL}$ can be a non-linear function and this will not hurt the interpretability, as our focus is solely on the activation behaviors of the neurons in the last layer of CBL. To force the last $k$ neurons in the $f_\textrm{CBL}$ learn the $k$ concepts, we maximize the similarity between $f_\textrm{CBL}(f_\textrm{LM}(x))$ and $S_c(x)$ for every $x$:
\begin{equation}
    \max_{\theta_1,\theta_2}\dfrac{1}{|\mathcal{D}|}\sum_{x\in\mathcal{D}}Sim\big(f_\textrm{CBL}(f_\textrm{LM}(x;\theta_1);\theta_2),S_c(x)\big),
\end{equation}
where $Sim: \mathbb{R}^k\times\mathbb{R}^k\rightarrow\mathbb{R}$ can be any similarity function, $\theta_1$ and $\theta_2$ are the parameters of the pretrained LM and the CBL respectively. 
%One can choose not to update $\theta_1$ to reduce training time, albeit at the expense of sacrificing some accuracy.

\paragraph{Learning the predictor:} After training the CBL, the $k$ neurons from the last layer of CBL learn the corresponding $k$ concepts. Let $A_N$ be the neuron activations from the last layer neurons of CBL $A_N(x)=f_\textrm{CBL}(f_\textrm{LM}(x))$, we set all the negative activations of $A_N(x)$ to zero through a ReLu function $A^+_N(x)=\textrm{ReLu}(A_N(x))$. We remove the negative activations as the negation of a concept introduces ambiguity (e.g., it is unclear whether the negative activations imply the absence of a concept or the negation of the semantic meaning of a concept). After obtaining $A^+_N$, we train a final linear layer with sparsity constraint to make predictions:
\begin{equation}
    \min_{W,b}\dfrac{1}{|\mathcal{D}|}\sum_{x,y\in\mathcal{D}}\mathcal{L}_\textrm{CE}(WA^+_N(x)+b, y)+\lambda R(W),
\end{equation}
where $W\in\mathbb{R}^{n\times k}$ is the weight matrix and $b\in\mathbb{R}^n$ is the bias vector of the final linear layer, $y$ is the label of $x$, and $R(W)=\alpha ||W||_1+(1-\alpha)\frac{1}{2}||W||_2^2$ is the elastic-net regularization, which is the combination of $\ell_1$ and $\ell_2$ penalty. Generally, a sparse final layer makes the CBM more interpretable. We will discuss the effect of sparsity in Section \ref{sec:experiment results}.
\section{Automatic Concept Correction}
\label{sec: language-model-guided concept-label intervention}
While ACS offers an efficient way to provide pseudo labels (concept scores), its correctness is dependent on the performance of the sentence embedding model. This introduces a limitation wherein the concept scores may not align with human reasoning, consequently impacting the learning of the CBL and introducing a trade-off in performance. Notably, this challenge is prevalent in recent CBM works that do not rely on human-assigned concept labels.

To address this challenge, we proposed Automatic Concept Correction (ACC), a technique leveraging the knowledge from ChatGPT to improve the quality of concept scores generated by ACS. As shown in our experiment (Table \ref{table:acc}), ACC can effectively boost the performance of CBM to a comparable level with black-box models.
 
Here, we describe the details of ACC. Recall that in Section \ref{sec: concept generation}, we generate the concept set $\mathcal{C}=\bigcup_{i=0}^{n-1} \mathcal{S}_i$ for dataset $\mathcal{D}$ with $n$ classes, where $\mathcal{S}_i$ is the concept subset for class $i$. We define the mapping $\mathcal{M}:c\rightarrow\{0,...,n-1\}$ which maps a concept $c\in\mathcal{C}$ to a class:
\begin{equation}
    \mathcal{M}(c)=
    \begin{cases}
        0\;\textrm{if}\; c\in\mathcal{S}_0 \\
        1\;\textrm{if}\; c\in\mathcal{S}_1 \\
        \vdots \\
        n-1\;\textrm{if}\; c\in\mathcal{S}_{n-1} \\
    \end{cases}
\end{equation}
For any text sample $x$ in $\mathcal{D}$, let $y$ be the class label of $x$ and $S_c(x)$ be the concept scores generated by sentence embedding model $\mathcal{E}$ as in Eq. \eqref{e:sc}. The key idea is to replace $S_c(x)$ with new concept scores $S_c^{\textrm{ACC}}(x)$, which are corrected by the ACC procedure. The new concept scores $S_c^{\textrm{ACC}}(x)$ are defined as follows:
\begin{equation}
        S_c^{\textrm{ACC}}(x)_i =
        \begin{cases}
            \mathcal{E}(c_i)\mathcal{E}(x),\;\textrm{if}\;\mathcal{E}(c_i)\mathcal{E}(x)>0, \mathcal{M}(c_i)=y \\
            0,\;\textrm{otherwise}
        \end{cases}
\end{equation}
where $S_c^{\textrm{ACC}}(x)_i$ is the $i$-th component of vector $S_c^{\textrm{ACC}}(x)$. ACC filters out the negative concept scores and forces every component of $S_c^{\textrm{ACC}}(x)$ to be zero when the corresponding concept $c_i$ and text sample $x$ belong to different classes. This is achievable because we prompt ChatGPT to generate the concept set for each class separately, thereby providing information about the association of concepts with their respective classes.

We utilize ACC to correct inaccurate concept scores before training the CBL, leading to a significant improvement in the accuracy of CB-LLM, which matches and, in certain cases, even surpasses those of finetuned black-box models. Further details on the accuracy of CB-LLM will be discussed in Section \ref{sec:accuracy of cbllm}. Unlike prior studies focusing on leveraging test-time intervention to correct the predictions of CBM, ACC occurs before the training of CBM and does not necessitate information about the testing set or any human knowledge. Additionally, our ACC strategy does not require any extra queries to ChatGPT and can be executed with almost zero time cost.
\section{Experiment results}
\label{sec:experiment results}

In this section, we evaluate our CB-LLM in terms of three crucial aspects: \emph{Accuracy}, \emph{Efficency}, and \emph{Faithfulness}. These aspects are pivotal as our goal is to ensure that CB-LLM achieves high accuracy with minimal additional cost while providing reasonable and human-understandable explanations.

\paragraph{Setup.} We conduct experiments on the standard text-classification benchmarks: 
\begin{itemize}
    \vspace{-5pt}
    \item \textbf{SST2 \cite{sst2}:} comprise 6920 training samples, 872 validation samples, and 1821 test samples of movie reviews with positive and negative classes.
    \vspace{-5pt}
    \item \textbf{Yelp Polarity (YelpP) \cite{yelpagnews}:} comprise 560,000 training samples and 38,000 test samples of Yelp reviews with positive and negative classes.
    \vspace{-5pt}
    \item \textbf{AGnews \cite{yelpagnews}:} comprise 120,000 training samples and 7,600 test samples of news articles with 4 classes.
    \vspace{-5pt}
    \item \textbf{DBpedia \cite{dbpedia}:} comprise 560,000 training samples and 70,000 test samples from DBpedia 2014 with 14 classes.
    \vspace{-5pt}
\end{itemize}
We generate $208$ concepts for SST2, $248$ concepts for YelpP, $216$ concepts for AGnews, and $476$ concepts for DBpedia. We use \textbf{\texttt{RoBERTa-base}} \cite{roberta} pretrained model with 768 output dimensions as the backbone for learning CB-LLM, and compared our CB-LLM with the finetuned \textbf{\texttt{RoBERTa-base}} (standard black-box model).

\subsection{Accuracy of CB-LLM}
\label{sec:accuracy of cbllm}
The test accuracy is shown in Table \ref{table:acc}. In general, our CB-LLMs demonstrate high accuracy across various datasets, including large ones such as YelpP and DBpedia. The CB-LLM implementation without ACC already achieves high accuracy: only a 1\textasciitilde5\% gap compared to the standard black-box model. This gap can be further eliminated: it can be seen that our ACC strategy, described in Section \ref{sec: language-model-guided concept-label intervention}, improves the accuracy significantly to the level of the standard black-box model. This indicates that ACC can effectively correct inaccurate concept scores and enhance learning on the given task. As for the effect of the sparse final layer, we do not observe a large performance drop after incorporating the sparsity constraint. In fact, CB-LLM with a sparse final layer, when combined with ACC, sometimes exhibits better accuracy than the counterpart with only ACC. This observation suggests that our ACC strategy works well with the sparsity constraint on the final layer. Overall, our CB-LLMs sometimes achieve higher accuracy than the standard black-box model (highlighted in blue in Table \ref{table:acc}), showcasing the possibility of building an interpretable model without incurring a trade-off in performance loss.

\begin{table}[!t]
\vspace{-5pt}
\caption {Test accuracy of CB-LLM. Our CB-LLMs achieve nearly identical performance as the standard black-box model after undergoing ACC. Numbers highlighted in blue indicate accuracy surpassing that of the standard black-box model.}
\label{table:acc}
\tabcolsep=0.12cm
\centering
\scriptsize
\begin{NiceTabular*}{\linewidth}{@{\extracolsep{\fill}} lcccc}[colortbl-like]
\toprule
    {Method} & \multicolumn{4}{c} {Dataset}\\
    \cmidrule(lr){2-5}
    {} & SST2 & YelpP & AGnews & DBpedia \\
    \midrule
    \textbf{Ours:} \\ 
    {CB-LLM} & $0.9138$ & $0.9358$ & $0.8989$ & $0.9828$ \\
    {CB-LLM w/ sparse final layer} & $0.9094$ & $0.9327$ & $0.8972$ & $0.9742$ \\
    {CB-LLM w/ ACC} & $\textcolor{blue}{0.9473}$ & $\textcolor{blue}{\bf{0.9805}}$ & $0.9462$ & $\textcolor{blue}{\bf{0.9925}}$ \\
    {CB-LLM w/ ACC \& sparse final layer} & $\textcolor{blue}{\bf{0.9478}}$ & $0.9803$ & $0.9467$ & $\textcolor{blue}{\bf{0.9925}}$ \\
    \midrule
    % \textbf{Ours (angle ACS):} \\ 
    % {CB-LLM} & $0.9336$ & $0.9648$ & $0.9036$ & $0.9837$ \\
    % {CB-LLM w/ sparse final layer} & $0.9325$ & $0.9643$ & $0.8992$ & $0.9772$ \\
    % {CB-LLM w/ ATI} & $0.9412$ & $0.9784$ & $0.9424$ & $\textcolor{blue}{0.9924}$ \\
    % {CB-LLM w/ ATI \& sparse final layer} & $0.9401$ & $0.9789$ & $0.9433$ & $\textcolor{blue}{0.9923}$ \\
    % \midrule
    \textbf{Baseline (standard black-box):} \\
    {Roberta-base finetuned} & $0.9418$ & $0.9803$ & $\bf{0.9478}$ & $0.9922$ \\
\bottomrule

\end{NiceTabular*}
\end{table}

\subsection{Efficiency of CB-LLM}
\label{sec:efficiency of cbllm}
The time cost of Automatic Concept Scoring (ACS) and finetuning language model is shown in Table \ref{table:efficiency}. Our ACS strategy takes about 1.6 hours on the largest YelpP and DBpedia dataset when using \textbf{\texttt{all-mpnet-base-v2}} as the sentence embedding model. The training time of CB-LLM is approximately equivalent to the time cost of finetuning the standard black-box model. These results indicate that we incur only a small overhead of time cost while significantly improving interpretability through the incorporation of a human-interpretable CBL and a final linear layer.

\begin{table}[!t]
\vspace{-5pt}
\caption {The time cost of ACS and learning CB-LLM. Training CB-LLM requires only a little additional time cost compared to finetuning the black-box language models.}
\label{table:efficiency}
\tabcolsep=0.015cm
\centering
\scriptsize
\begin{tabular*}{\linewidth} {@{\extracolsep{\fill}} lcccc}
\toprule
    {Time cost (hours)}  & \multicolumn{4}{c} {Dataset}\\
    \cmidrule(lr){2-5}
    {} & SST2 & YelpP & AGnews & DBpedia \\
    \midrule
    \multicolumn{3}{l} {\textbf{Automatic Concept Scoring (ACS):}} \\
    {mpnet ACS} & $0.0024$ & $1.6172$ & $0.2455$ & $1.6578$ \\
    \midrule
    % {Automatic Training Intervention (ATI)} & $0.0005$ & $0.0460$ & $0.0063$ & $0.0542$\\
    % \midrule
    \textbf{Finetuning model:} \\
    {CB-LLM} & $0.0984$ & $8.9733$ & $2.0270$ & $9.1800$\\
    {Standard black-box} & $ 0.0289$ & $8.9679$ & $1.3535$ & $9.1996$\\
\bottomrule

\end{tabular*}
\vspace{-5pt}
\end{table}

\subsection{Faithfulness of CB-LLM}
\label{sec:faithfulness of cbllm}
It is important for an interpretable model to make predictions based on human-understandable and faithful logic. Hence, in this section, we introduce our faithfulness evaluation design and present the results obtained through large-scale human evaluation for our CB-LLM.

We define two metrics to evaluate the faithfulness :
\begin{enumerate}
    \item \textbf{Activation Faithfulness:} This evaluates if the activations of neurons in CBL align with the corresponding concepts they have learned. For a given neuron in CBL, its activation over each text sample in the dataset can be extracted. Subsequently, \emph{Activation Faithfulness} can be evaluated by manually inspecting whether the highly activated samples are related to the concept represented by the given neuron.
    \item \textbf{Contribution faithfulness:} This evaluates if the activation of neurons in CBL makes reasonable contributions to the final predictions. We first define the contribution of a neuron in CBL. For any text sample $x$ from dataset $\mathcal{D}$ with $n$ classes, the contribution for neuron $j$ to class $i$ is denoted as $W_{ij}{A_N}^+(x)_j$, where $W$ is the weight matrix from the final linear layer, ${A_N}^+(x)$ is the non-negative activations from CBL, $i\in\{1,...,n\}$, $j\in\{1,...,k\}$, $n$ is the number of classes, and $k$ is the number of neurons in CBL. This contribution score directly describes how each neuron in CBL influences the final predictions. For a text sample that is correctly classified, we extract the neurons in CBL with high contributions to the prediction. Subsequently, \emph{Contribution Faithfulness} can be evaluated by manually inspecting whether the concepts represented by the highly contributed neurons are related to the given text sample and are reasonable for making the correct prediction.
\end{enumerate}

\paragraph{Human evaluation design.} We perform the human evaluation through Amazon Mechanical Turk (MTurk) to study the faithfulness of our CB-LLM. Based on the above metrics, we design two tasks for human evaluation:
\begin{enumerate}
    \item \textbf{Task 1: Activation Faithfulness.} In this task, workers will be presented with a neuron concept alongside the corresponding top $k$ highly activated text samples. Workers need to provide a rating ranging from 1 (strongly disagree) to 5 (strongly agree) based on the agreement observed between the neuron concept and the top $k$ highly activated samples.
    \item \textbf{Task 2: Contribution Faithfulness.} In this task, workers will be presented with explanations from two models for a text sample. The explanations are generated by showing the top $r$ neuron concepts with the highest contribution to the prediction. Workers need to compare which model's explanations are better. 
    %and select an option from "model 1 is clearly better", "model 1 is slightly better", "equally good", "model 2 is slightly better", and "model 2 is clearly better".
\end{enumerate}
We conduct human evaluations for Task 1 and Task 2 to compare our CB-LLMs with the \emph{Random baseline}. The random baseline is generated by the following rules: For Task 1, the highly activated text samples are randomly selected. For Task 2, the explanations are randomly selected from the same concept set. To ensure more reliable results, each question in the tasks mentioned above is evaluated three times by different workers.

We also perform an additional large-scale human evaluation to further verify the human evaluation results for CB-LLMs. More details about the survey design and interface can be found in Appendix \ref{sec:interface}.

\begin{table}[t!]
\vspace{-5pt}
\caption {The human evaluation results for task 1 --- Activation Faithfulness. Workers give higher ratings to our CB-LLM w/ ACC, suggesting that the neurons in our CB-LLM w/ ACC are more interpretable than the neurons with random activations.}
\label{table:task1}
\centering
\tabcolsep=0.05cm
\scriptsize
\begin{NiceTabular*}{\linewidth} {@{\extracolsep{\fill}} lcccc|c}
\toprule
    \multicolumn{1}{l} {Task 1 --- Activation Faithfulness} & \multicolumn{4}{c} {Dataset} & Average \\
    \cmidrule(lr){2-5}
    \multicolumn{1}{l} {Method} & SST2 & YelpP & AGnews & DBpedia \\
    \midrule
    \multicolumn{5}{l|} {\textbf{Human evaluation:}} \\ 
    \multicolumn{1}{l} {CB-LLM w/ ACC (Ours)} & $\bf{4.07}$ & $\bf{4.00}$ & $\bf{4.00}$ & $\bf{4.07}$ & $\bf{4.03}$ \\
    % \multicolumn{1}{l} {(angle) CB-LLM w/ ATI} & $3.73$ & $3.33$ & $\bf{4.27}$ & $\bf{4.40}$ & $3.93$ \\
    \multicolumn{1}{l} {Random (Baseline)} & $3.40$ & $3.53$ & $2.86$ & $2.80$ & $3.15$ \\
    \midrule
    \multicolumn{5}{l|} {\textbf{Large-scale human evaluation:}} \\
    \multicolumn{1}{l} {CB-LLM w/ ACC (Ours)} & $\bf{3.50}$ & $\bf{4.03}$ & $\bf{4.23}$ & $\bf{3.90}$ & $\bf{3.92}$ \\
    % \multicolumn{1}{l} {(angle) CB-LLM w/ ATI} & $3.40$ & $3.73$ & $3.77$ & $3.80$ & $3.68$ \\
    % \midrule
    % \multicolumn{5}{l|} {\textbf{Baseline:}} \\
    \multicolumn{1}{l} {Random (Baseline)} & $3.03$ & $3.20$ & $2.97$ & $3.13$ & $3.08$ \\
\bottomrule
\end{NiceTabular*}
\vspace{-5pt}
\end{table}
\begin{figure}[!t]
\centering
\includegraphics[width=0.48\textwidth]{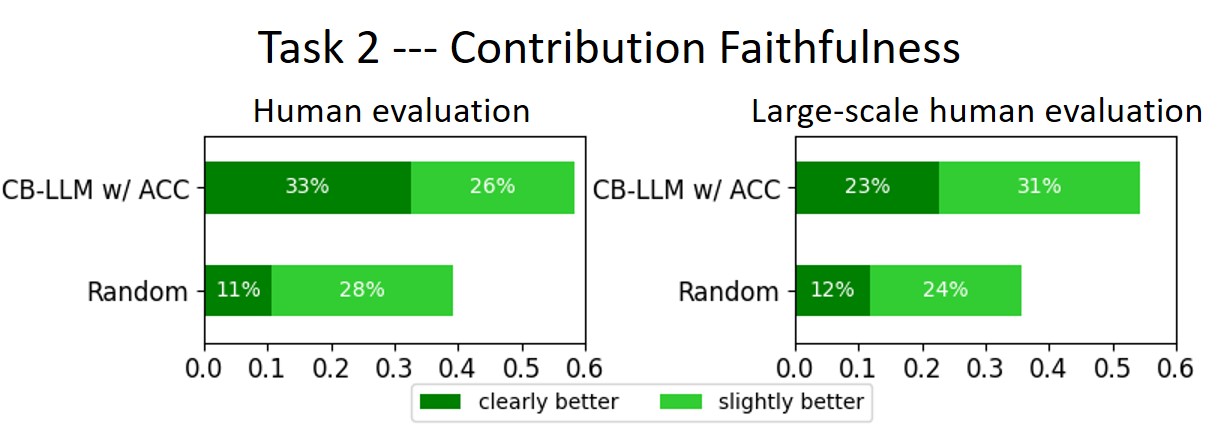}
\vspace{-20pt}
\caption{The human evaluation results for task 2 --- Contribution Faithfulness. Workers prefer the explanations generated by CB-LLM w/ ACC more than the random explanations.}
\label{fig:task2}
\vspace{-5pt}
\end{figure}

\subsubsection{Results of human evaluation}
\label{sec: large scale}
The results of task 1 (Activation Faithfulness) are shown in Table \ref{table:task1}. Our CB-LLMs w/ ACC constantly achieve higher ratings than the random baseline. This suggests that the neurons in our CB-LLMs w/ ACC are more interpretable than the neurons with random activations.

The results of task 2 (Contribution Faithfulness) are shown in Figure \ref{fig:task2}. Workers consistently express a preference for our CB-LLM w/ ACC over the random baseline. This suggests that the explanations generated by our CB-LLM w/ ACC are better than randomly selected explanations.

\subsubsection{Ablation study}
\label{sec: small scale}
\begin{figure}[!t]
\centering
\includegraphics[width=0.48\textwidth]{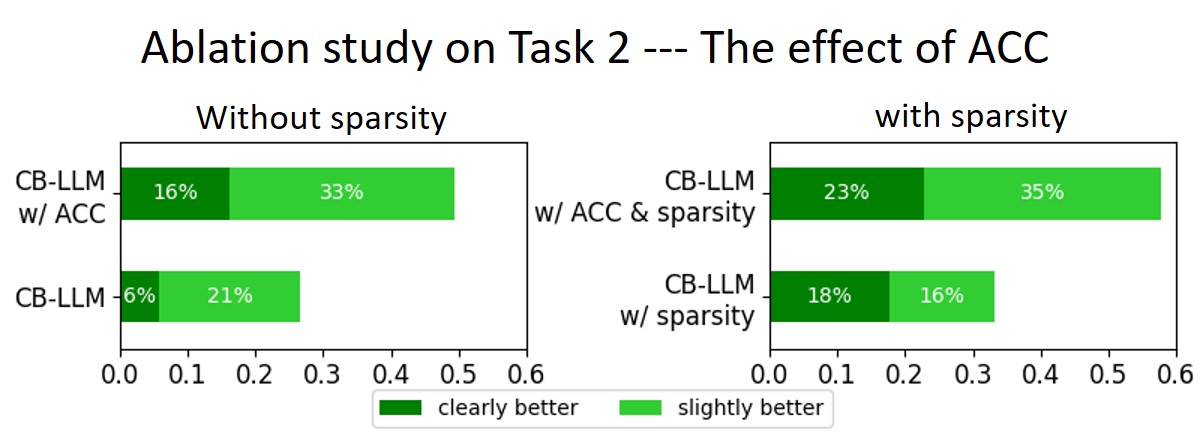}
\vspace{-20pt}
\caption{Ablation study on Automatic Concept Correction (ACC). Workers favor the explanations provided by the CB-LLMs with ACC.}
\label{fig:intervention}
\vspace{-5pt}
\end{figure}

\begin{figure}[!t]
\centering
\includegraphics[width=0.48\textwidth]{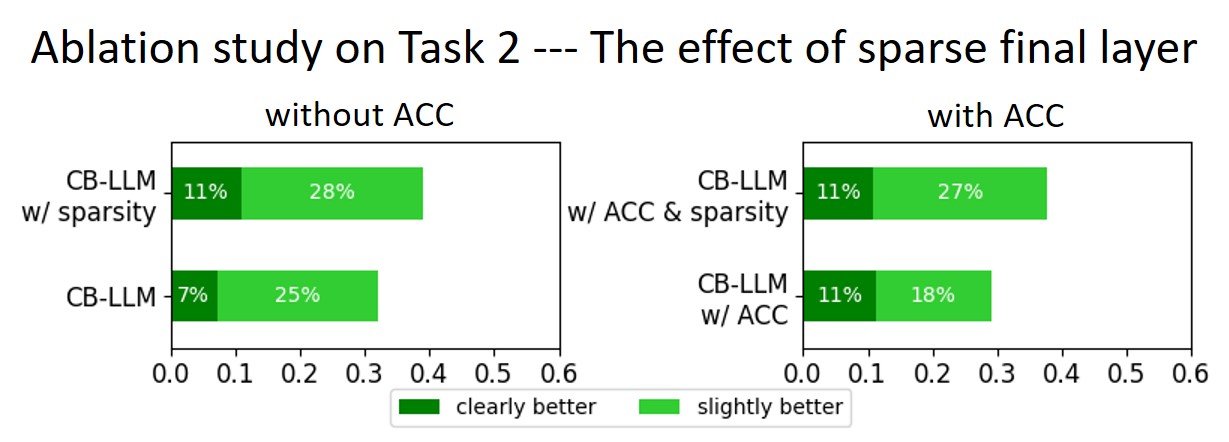}
\vspace{-20pt}
\caption{Ablation study on the sparsity. Workers demonstrate only a marginal preference for explanations provided by the CB-LLMs with a sparse final layer.}
\label{fig:sparsity}
\vspace{-5pt}
\end{figure}

We conduct an ablation study to evaluate the impact of ACC and sparse final layer on faithfulness through task 2. Figure \ref{fig:intervention} compares the CB-LLMs with ACC and the ones without ACC. Workers consistently prefer the explanations from CB-LLM with ACC, irrespective of using a sparse final layer or not. This indicates that our ACC strategy not only improves accuracy but also enhances the quality of explanations. Figure \ref{fig:sparsity} compares the CB-LLMs w/ sparse final layer and the ones without. Workers exhibit little preference after the application of a sparse final layer, suggesting that sparsity might offer marginal help for the interpretability of CB-LLM. More details about the ablation study can be found in Appendix \ref{sec:detailed ablation}.

\section{Case study: Concept Unlearning}
\label{sec:case study}

\begin{table*}[!t]
\vspace{-10pt}
\caption {The two neurons from CB-LLM w/ ACC and their corresponding highly activated samples.}
\label{table:nueron main}
\centering
\fontsize{7.7}{8.5}\selectfont
\begin{tabular*}{\linewidth} {|p{2.6cm}|p{13.655cm}|}
\hline
    Neuron & Highly activated samples \\
    \hline
    \parbox{\hsize}{\textbf{(AGnews) Neuron \#16:}\\human rights violations and advocacy.} & \parbox{\hsize}{
    \begin{enumerate}
        \vspace{-5pt}
        \item US soldier convicted of torture in Iraq A US military intelligence soldier in Iraq has been sentenced to 8 months in prison for taking part in torturing detainees in Abu Ghraib prison.
        \vspace{-5pt}
        \item Pinochet is ordered to stand trial for murder Augusto Pinochet, the former Chilean dictator, was ordered under house arrest yesterday, charged with kidnapping and murder dating back to his 17-year rule.
        \vspace{-5pt}
        \item Trial Date Set for Soldier at Abu Ghraib (AP) AP - A military judge ordered a U.S. Army reservist on Friday to stand trial Jan. 7 in Baghdad for allegedly abusing Iraq inmates at the Abu Ghraib prison outside Baghdad.
        \vspace{-5pt}
        \item Afghan court convicts US trio of torture KABUL, Afghanistan -- Three Americans -- led by a former Green Beret who boasted he had Pentagon support -- were found guilty yesterday of torturing Afghans in a private jail and were sentenced to prison.
        \vspace{-5pt}
        \item Soldier to Plead Guilty in Iraq Abuse Case (AP) AP - An Army reservist charged with abusing Iraqi prisoners plans to plead guilty at a court martial to four counts arising from the Abu Ghraib prison abuse scandal in a plea deal in which eight other counts will be dropped, his lawyer has said.
        \vspace{-5pt}
    \end{enumerate}
    } \\
    \hline
    \parbox{\hsize}{\textbf{(DBpedia) Neuron \#71:}\\the artist's born date.} & \parbox{\hsize}{
    \begin{enumerate}
        \vspace{-5pt}
        \item Joanna Taylor (born 24 July 1978) is an English actress and former model.
        \vspace{-5pt}
        \item Jody Miller (born November 29 1941) is an American country music singer. Born as Myrna Joy Miller she was born in Phoenix Arizona and raised in Oklahoma.
        \vspace{-5pt}
        \item Priscilla Mitchell (born September 18 1941 in Marietta Georgia) was an American country music singer.
        \vspace{-5pt}
        \item Geoffrey Davies (born 15 December 1942 Leeds West Riding of Yorkshire) is a British actor.
        \vspace{-5pt}
        \item He was born in Asunción Paraguay on March 27 1950. Son of Carmen Emategui and Rodolfo Barreto.
        \vspace{-5pt}
    \end{enumerate}
    } \\
\hline
\end{tabular*}
\end{table*}
\begin{table*}[!t]
\vspace{-10pt}
\caption {The explanations generated by CB-LLM w/ ACC for two text samples.}
\label{table:explanation main}
\centering
\fontsize{7.7}{8.5}\selectfont
\begin{tabular*}{\linewidth} {|p{11.35cm}|p{4.908cm}|}
\hline
    Sample & Explanations \\
    \hline
    \parbox{\hsize}{\textbf{(SST2) Sample \#330:}\\occasionally funny , always very colorful and enjoyably overblown in the traditional almodóvar style .} & \parbox{\hsize}{
    \begin{enumerate}
        \vspace{-5pt}
        \item Charming characters.
        \vspace{-5pt}
        \item Clever and unexpected humor.
        \vspace{-5pt}
        \item Stunning and exotic locations.
        \vspace{-5pt}
        \item Stellar and diverse ensemble cast.
        \vspace{-5pt}
        \item Unique and well-developed characters.
        \vspace{-5pt}
    \end{enumerate}
    } \\
    \hline
    \parbox{\hsize}{\textbf{(YelpP) Sample \#34857:}\\This place has something for everyone.  My wife and I started going there out of convenience before attending a movie at the South Pointe.  But then we continued going back because we liked the food and the staff is very helpful.  This most recent visit I had sushi for the first time and it was very good - and reasonably priced.  We have company coming and are going to make it one of our stops on their visit.} & \parbox{\hsize}{
    \begin{enumerate}
        \vspace{-5pt}
        \item Welcoming and friendly staff.
        \vspace{-5pt}
        \item Clean and inviting ambiance.
        \vspace{-5pt}
        \item Amazing flavors.
        \vspace{-5pt}
        \item Great warranty and support.
        \vspace{-5pt}
        \item Delicious food.
        \vspace{-5pt}
    \end{enumerate}
    } \\
\hline
\end{tabular*}
\vspace{-5pt}
\end{table*}

In this section, we provide use cases to demonstrate how to leverage the interpretability of our CB-LLM in practice.

\textbf{Concept Unlearning} refers to forcing the model to forget a certain concept. In some situations, there might be specific reasons to deactivate the influence of a particular concept on the final prediction. With the interpretable structure of our CB-LLM, we can easily unlearn a concept by manually deactivating a specific neuron in the CBL or removing all the weights connected to this neuron in the final linear layer.

\begin{figure}[!t]
\centering
\includegraphics[width=0.48\textwidth]{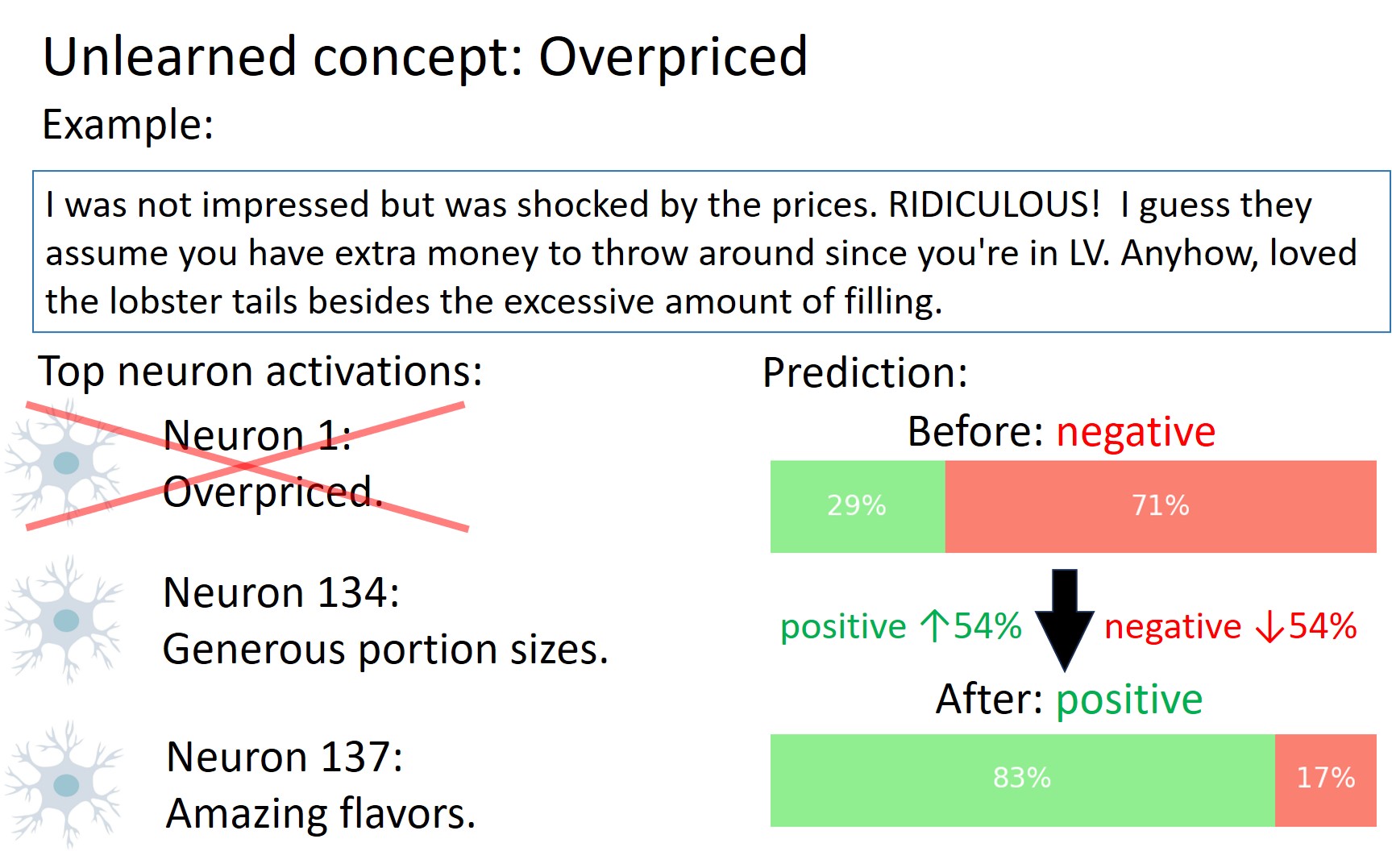}
\vspace{-20pt}
\caption{An example of concept unlearning. This example is initially classified as negative due to the customer complaining about the high price, despite the lobster tails being great. After unlearning the concept "Overpriced", the concepts "Amazing flavors" and "Generous portion sizes" dominate the prediction, resulting in a positive prediction.}
\vspace{-8pt}
\label{fig:unlearning}
\end{figure}

Figure \ref{fig:unlearning} presents an example of unlearning the concept of "overpriced". In practice, we might consider removing the concept of "overpriced" from Yelp reviews due to subjectivity or geographical reasons (as the standard of overpricing varies across individuals and locations). This adjustment can encourage CB-LLM to prioritize the evaluation of product quality. After unlearning the concept of "overpriced," the predictions for 2726 samples in the test set changed from negative to positive. Subsequently, we employed \textbf{\texttt{bart-large-mnli}}, an NLI model, to assess whether these 2726 samples indeed contain the concept of "overpriced". Our findings reveal that 2162 out of the 2726 samples strongly entail "overpriced", accounting for 79\%. This suggests that most of the samples now predicting positive were initially classified as negative due to the presence of the "overpriced" concept.

Based on the above case study, we believe our CB-LLM has great potential to facilitate human intervention such as Concept Unlearning for enhancing fairness, as users can easily remove biased, subjective, or unfair elements that could distort the predictions. More examples of Concept Unlearning can be found in Appendix \ref{sec:more examples}.

\section{Visulization of neurons and explanations}
\label{sec:visulization}
We provide some visualizations of neurons in our CB-LLM w/ ACC in Table \ref{table:nueron main}. The neurons are displayed along with their corresponding concepts and the top 5 highly activated samples in the dataset. For a given text sample, we also show the 5 explanations generated by CB-LLM w/ ACC in Table \ref{table:explanation main}. The explanations are generated by selecting the top 5 concepts with the highest contribution to the prediction. More visualizations of neurons and explanations can be found in Appendix \ref{sec:neuron visualization} and \ref{sec:explanations} respectively.
\section{Conclusion}
\label{sec:conclusion}
In this work, we introduced CB-LLM, the first CBM that scales to large text classification benchmarks. Our CB-LLM is fully automatic, training-efficient, and achieves accuracy comparable to, and sometimes surpassing, black-box language models while providing faithful interpretability. 
%In future work, we consider the idea of developing a CBM for autoregressive generative language models, such as GPT, to be an interesting direction, which is not limited to text classification tasks and holds potential for a broader range of applications.

\section*{Broader impact}
CB-LLM represents a notable advancement in the realm of interpretable language models. As demonstrated in Section \ref{sec:case study}, CB-LLM allows human intervention to identify and remove biased, subjective, or unfair elements that can distort the predictions. We believe that this feature could positively contribute to fairness and transparency in the development of Large Language Models.

\section*{Acknowledgements}
This work is supported in part by National Science Foundation (NSF) awards CNS-1730158, ACI-1540112, ACI-1541349, OAC-1826967, OAC-2112167, CNS-2100237, CNS-2120019, the University of California Office of the President, and the University of California San Diego's California Institute for Telecommunications and Information Technology/Qualcomm Institute. Thanks to CENIC for the 100Gbps networks. C. Sun, T. Oikarinen and T.-W. Weng are supported by National Science Foundation under Grant No. 2107189 and 2313105. T.-W. Weng also thanks the Hellman Fellowship for providing research support.

\bibliography{ref}
\bibliographystyle{icml2024}

%%%%%%%%%%%%%%%%%%%%%%%%%%%%%%%%%%%%%%%%%%%%%%%%%%%%%%%%%%%%%%%%%%%%%%%%%%%%%%%
%%%%%%%%%%%%%%%%%%%%%%%%%%%%%%%%%%%%%%%%%%%%%%%%%%%%%%%%%%%%%%%%%%%%%%%%%%%%%%%
% APPENDIX
%%%%%%%%%%%%%%%%%%%%%%%%%%%%%%%%%%%%%%%%%%%%%%%%%%%%%%%%%%%%%%%%%%%%%%%%%%%%%%%
%%%%%%%%%%%%%%%%%%%%%%%%%%%%%%%%%%%%%%%%%%%%%%%%%%%%%%%%%%%%%%%%%%%%%%%%%%%%%%%
\newpage
\appendix
\onecolumn

\section{Appendix}

\subsection{MTurk survey design and interface}
\label{sec:interface}
We perform the human evaluation through Amazon Mechanical Turk (MTurk). The details of two tasks we design are as follows:
\begin{enumerate}
    \item \textbf{Task 1 --- Activation Faithfulness:} In this task, workers will be presented with a neuron concept alongside the corresponding top 5 highly activated text samples. Workers need to provide a rating ranging from 1 (strongly disagree) to 5 (strongly agree) based on the agreement observed between the neuron concept and the top 5 highly activated samples.
    \item \textbf{Task 2 --- Contribution Faithfulness.} In this task, workers will be presented with explanations from two models for a text sample. The explanations are generated by showing the top 5 neuron concepts with the highest contribution to the prediction. Workers need to compare which model's explanations are better and select an option from "model 1 is clearly better", "model 1 is slightly better", "equally good", "model 2 is slightly better", and "model 2 is clearly better".
\end{enumerate}
We did human evaluations on MTurk for Task 1 and Task 2 as mentioned in Section \ref{sec:faithfulness of cbllm}. The details are as follows:
\begin{itemize}
    \item \textbf{Human evaluation:} We evaluate the following 5 models:
    \begin{itemize}
        \item CB-LLM (Vanilla)
        \item CB-LLM w/ ACC
        \item CB-LLM w/ sparse final layer
        \item CB-LLM w/ ACC \& sparse final layer
        \item \emph{Random baseline}: For Task 1, the highly activated text samples are randomly selected. For Task 2, the explanations are randomly selected from the same concept set.
    \end{itemize} 
    For task 1, we evaluate each model's 5 most highly activated neuron concepts across each dataset. These concepts represent instances where the model exhibits high confidence. For task 2, we evaluate 5 random samples for every dataset.
\end{itemize}
We also did a large-scale human evaluation to verify the performance of our CB-LLM w/ ACC. The details are as follows:
\begin{itemize}
    \item \textbf{Large-scale human evaluation:} We evaluate the following 2 models:
    \begin{itemize}
        \item CB-LLM w/ ACC
        \item \emph{Random baseline}: For Task 1, the highly activated text samples are randomly selected. For Task 2, the explanations are randomly selected from the same concept set.
    \end{itemize}
    For Task 1, we evaluate the 10 most highly activated neuron concepts for each model across each dataset. For task 2, we evaluate 40 random samples (20 per class) from SST2, 40 random samples (20 per class) from YelpP, 40 random samples (10 per class) from AGnews, and 70 random samples (5 per class) from DBpedia.
\end{itemize}
To ensure more reliable results, each question in the tasks mentioned above is evaluated three times by different workers. 

\begin{figure}[H]
\centering
\includegraphics[width=1\textwidth]{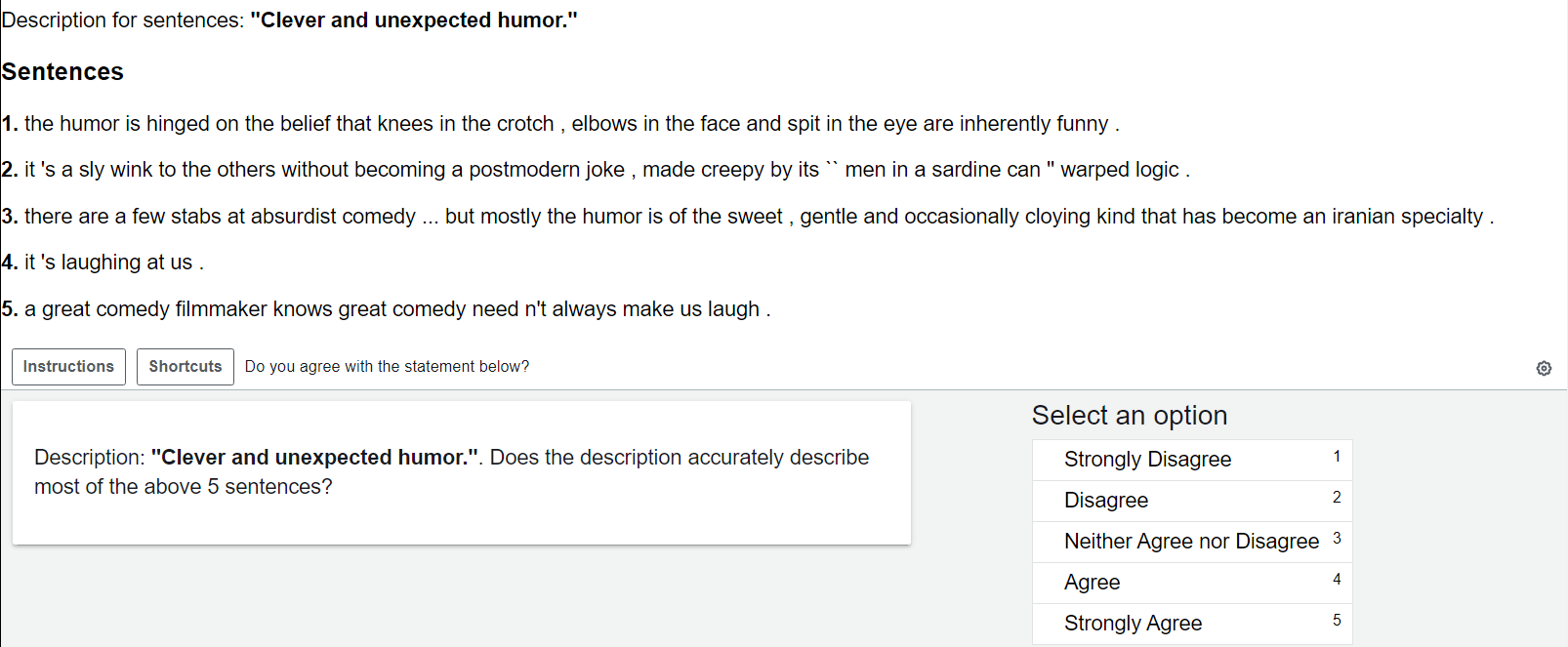}
\vspace{-20pt}
\caption{The interface for task 1 --- Activation faithfulness.}
\label{fig:task1 screen}
\end{figure}

\begin{figure}[H]
\centering
\includegraphics[width=1\textwidth]{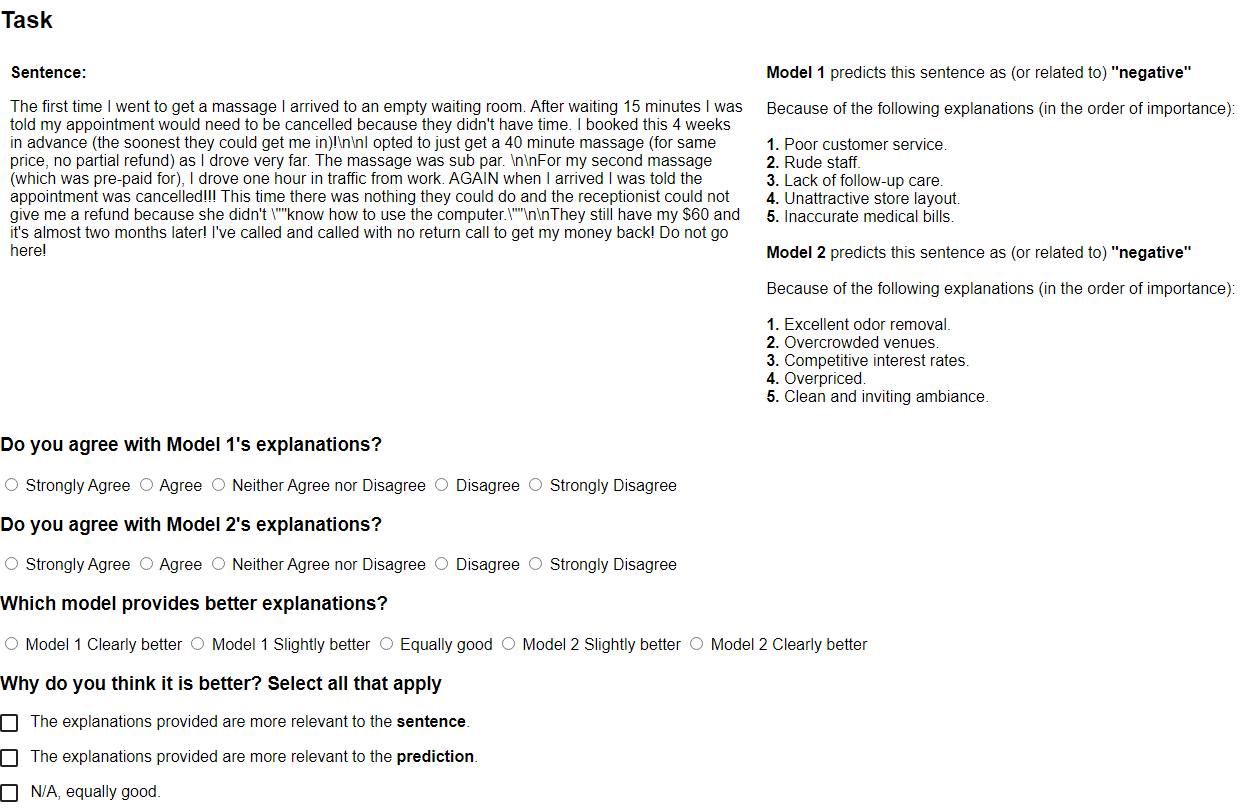}
\vspace{-20pt}
\caption{The interface for task 2 --- Contribution faithfulness.}
\label{fig:task2 screen}
\end{figure}

The survey interface for task 1 and task 2 is shown in Figure \ref{fig:task1 screen} and Figure \ref{fig:task2 screen} respectively. In task 2, workers are also asked to provide ratings for each model, similar to task 1. These ratings are utilized to filter out inconsistent results. The following logic is employed for filtering:
\begin{itemize}
    \item If workers indicate that model 1 is slightly or clearly better than model 2, the rating of model 1 must be no lower than the rating of model 2, and vice versa.
    \item If workers select "equally good," the two models must have the same rating.
\end{itemize}
\clearpage

\subsection{More details for ablation study}
We also compare the Vanilla CB-LLM without ACC or a sparse final layer with the random baseline. Vanilla CB-LLM still provides more favorable explanations compared to the random explanations.
\label{sec:detailed ablation}
\begin{figure}[H]
\centering
\includegraphics[width=0.7\textwidth]{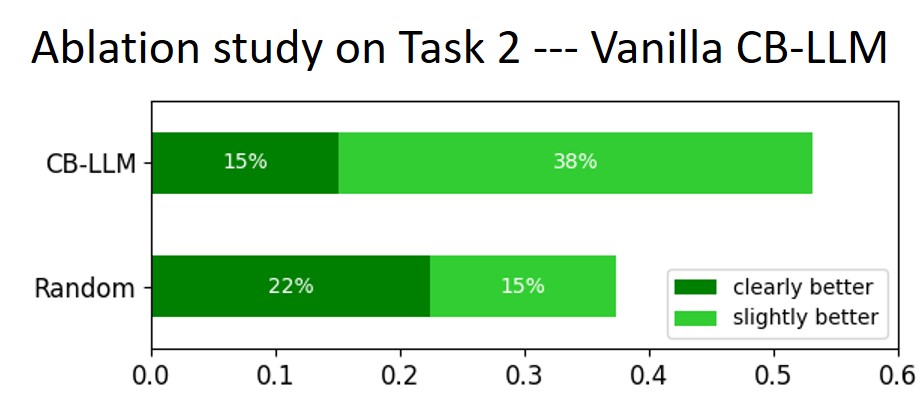}
\vspace{-10pt}
\caption{The evaluation of vanilla CB-LLM on task 2. Workers still prefer the explanations from Vanilla CB-LLM more than the explanations from random baseline.}
\label{fig:vanilla}
\end{figure}

\subsection{More examples for Concept Unlearning}
\label{sec:more examples}
\begin{figure}[H]
\centering
\includegraphics[width=0.8\textwidth]{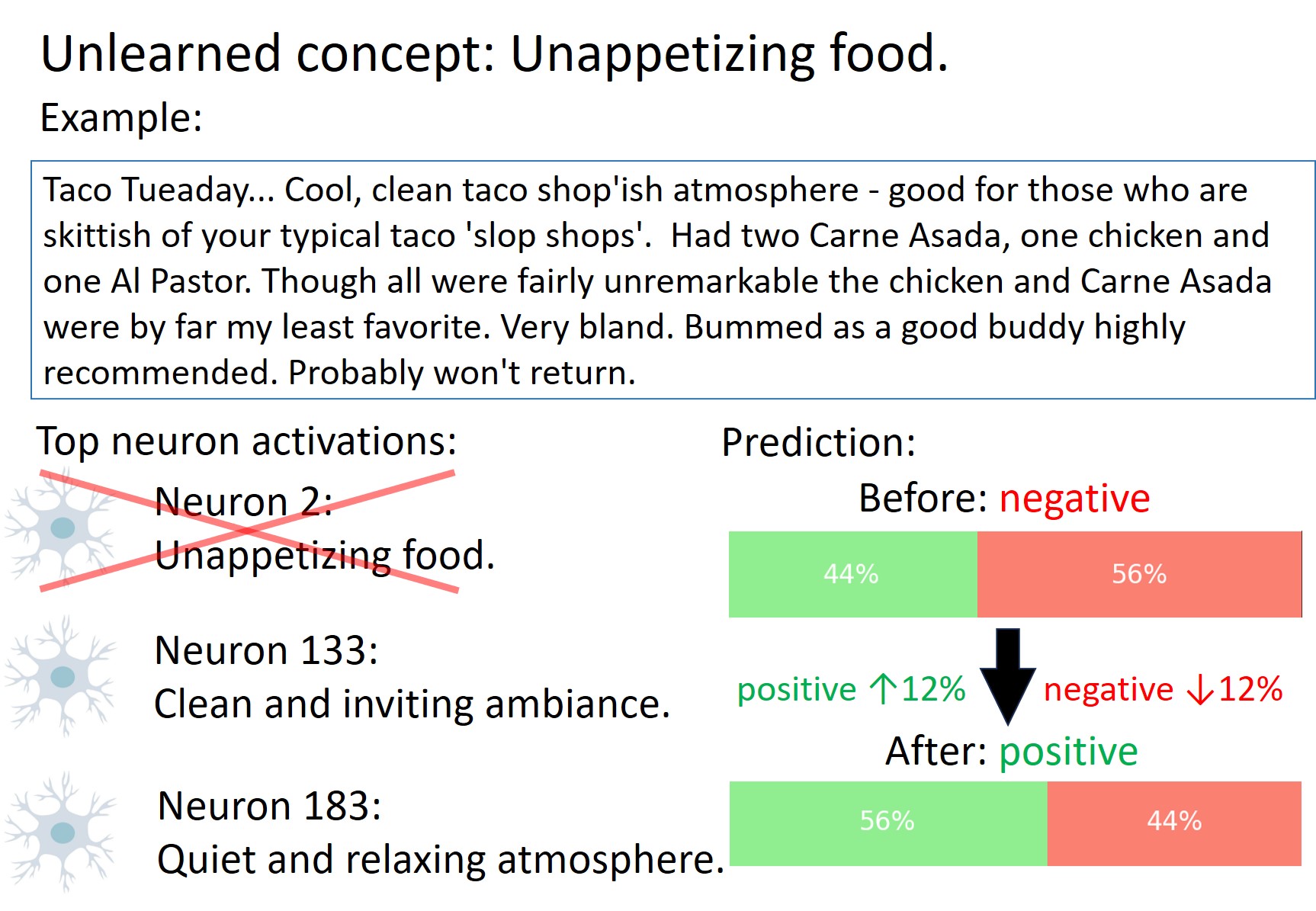}
\vspace{-10pt}
\caption{Another example of concept unlearning. This example is initially classified as negative due to the customer complaining about the bland food, despite the cool and clean atmosphere. After unlearning the concept "Unappetizing food" the concepts "Clean and inviting ambiance" and "quiet and relaxing atmosphere" dominate the prediction, resulting in a positive prediction.}
\label{fig:unlearning2}
\end{figure}

Figure \ref{fig:unlearning2} demonstrates another example of Concept Unlearning. The concept "Unappetizing food" is unlearned. After unlearning, the predictions of 370 samples changed from negative to positive, with 313 of them (85\%) strongly entailing "Unappetizing food". This suggests that most of the samples now predicting positive were initially classified as negative due to the presence of the "Unappetizing food" concept.

\clearpage

\subsection{Visualization of neurons in CB-LLM}
\label{sec:neuron visualization}
In this section, we provide more visualizations of the neurons in our CB-LLM. We select 3 neurons that have the highest activations across samples for each dataset.
\LTcapwidth=0.98\textwidth
\begin{longtable}{|l|p{4.5cm}|p{10cm}|}
\caption {The neurons of CB-LLM w/ ACC and corresponding highly activated samples for each dataset. We show the top 3 neurons with the largest activations for each dataset.}
\vspace{-10pt}
\label{table:neuron} \\
\hline
    Dataset & Neuron & Highly activated samples \\
    \hline
    {SST2} & \parbox{\hsize}{Neuron 184:\\Clever and unexpected humor.} & \parbox{\hsize}{
    \begin{enumerate}
        \item the humor is hinged on the belief that knees in the crotch , elbows in the face and spit in the eye are inherently funny .
        \item it 's a sly wink to the others without becoming a postmodern joke , made creepy by its `` men in a sardine can '' warped logic .
        \item there are a few stabs at absurdist comedy ... but mostly the humor is of the sweet , gentle and occasionally cloying kind that has become an iranian specialty .
        \item it 's laughing at us .
        \item a great comedy filmmaker knows great comedy need n't always make us laugh .
    \end{enumerate}
    } \\
    \hline
    {SST2} & \parbox{\hsize}{Neuron 170:\\Great chemistry between actors.} & \parbox{\hsize}{
    \begin{enumerate}
        \item when your leading ladies are a couple of screen-eating dominatrixes like goldie hawn and susan sarandon at their raunchy best , even hokum goes down easily .
        \item binoche and magimel are perfect in these roles .
        \item hugh grant and sandra bullock are two such likeable actors .
        \item interacting eyeball-to-eyeball and toe-to-toe , hopkins and norton are a winning combination -- but fiennes steals ` red dragon ' right from under their noses .
        \item without resorting to hyperbole , i can state that kissing jessica stein may be the best same-sex romance i have seen .
    \end{enumerate}
    } \\
    \hline
    {SST2} & \parbox{\hsize}{Neuron 34:\\Lack of humor or wit.} & \parbox{\hsize}{
    \begin{enumerate}
        \item frenetic but not really funny .
        \item but here 's the real damn : it is n't funny , either .
        \item francophiles will snicker knowingly and you 'll want to slap them .
        \item beyond a handful of mildly amusing lines ... there just is n't much to laugh at .
        \item do not , under any circumstances , consider taking a child younger than middle school age to this wallow in crude humor ."
    \end{enumerate}
    } \\
    \hline
    {YelpP} & \parbox{\hsize}{Neuron 184:\\Good breakfast options.} & \parbox{\hsize}{
    \begin{enumerate}
        \item I'm obsessed with the breakfast here. There's a huge smorgasbord of options to choose from on the brekkie menu, and the hardest part is actually picking something to order because they all sound so good! I couldn't resist ordering the eggs benedicto. What a cute twist on your typical eggs benedict dish! The eggs were perfectly poached on toasty slabs of english muffin and accented with the rich and savory sundried tomato hollandaise. The bits of candied prosciutto added a nice meatiness to the benedict without making it too heavy. And while I don't normally reach for mixed greens for breakfast.... I did like it in this dish because my usual gripe with eggs benedict is that there's just wayyy too much going on. But the greens were a light alternative that kinda balanced everything out in a way that potatoes don't do it for me. I also picked up the horchata latte. I'm a huge fan of horchata (which is pretty hard to find in Hawaii where I'm from) and a coffee lover, so this was a must try for me! It's totally sweet, creamy, and probably chock full of calories, but worth every single tasty sip. If you're not feeling in a benedicto mood, that's OK because there's a ton of other food options to choose from. All of which resemble your standard breakfast fare, with a little bit of a twist. Mexican, southern, classic american breakfasts... You name it. If I had more stomach room and a little more time in Madison, I'd wanna try a little bit of every dish on the menu. One of each, please!
        \item Half order of Mashed Potatoes Omelet and an ice tea is how everyone should start their day!
        \item Quite delicious for brunch. I am not normally a sweet breakfast food person, however the buckwheat waffle with a mimosa seems to be a perfect combination.
        \item The breakfast took a long time but when it finally did it was good! But a little pricey for eggs and bacon!
        \item Great breakfast.
    \end{enumerate}
    } \\
    \hline
    {YelpP} & \parbox{\hsize}{Neuron 159:\\Engaging performances.} & \parbox{\hsize}{
    \begin{enumerate}
        \item I saw LOVE yesterday, my first Las Vegas show. It was mind-bogglingly fantastic. I was totally swept away and mesmerized for over two hours. The sheer creativity, imagination, music, engineering, intricate choreography left me in a state of deep admiration for the entire effort. It was superb beyond words. See it before you die.
        \item If you're a huge Beatles fan, you will love this show. If you're a huge Cirque du Soleil fan, you might feel a lil' bit disappointed? But I guarantee this, you will definitely appreciate the artistic value of the show and what it's goal was..and that was to pay homage to one of the most influential bands in the history of music. Since I have been a life long Beatles fan, I was very curious as what I should expect? And then my wife literally said, """Let It Be!""" , and I did...I just relaxed and let go of any expectations from any other show that had seen in the past. Once the music began, I was knocked into the back of my seat. The audio and visual presentation is awesome. In addition to the theaters dynamic audio system, you have high end audio speakers that are embedded in the headrest of your very comfortable theater chairs, plus the seats that are in front of you have the speakers directed towards you as well!  The main body of the show starts off very solemn,  and as the crescendo builds....it EXPLODES on to the scene with a re-mastered version of """Get Back""". With performers  dancing and running around the middle of the stage, skaters skating, and acrobats literally falling and flying from the sky.....Whew!
        \item this show was great!! if you love fire and acrobatic stuff you will love this show!! its good for families as well. this was the 3rd cirque du soleii show they never dissapoint me. the set was awesome and costumes!
        \item Great show!   Great acts!   Wally Eastwood was awesome and funny.   Had 2 finalists from America got talent show.   I would see it again - great for all ages!   Acrobatics were cool.   The magic show was ok but still good to see as 1 of the many acts.   Wally Eastwood is on YouTube.  I highly recommend this show. The only missing star is for people expecting great props and scenery but for the great show,  you wouldn't care.
        \item what a really fun show!  It was really well paced and had a great selection of Beatles music.  The story line runs you through the decades.  The use of multi-media is really great and any seat in the house would be an incredible show.  it's a circular stage so there is stuff going on everywhere - it's hard to know where to look!! I thought the Cirque stuff was a little less insane than some of their other shows.  don't get me wrong - stunning and fun to watch but it didn't seem as over the top/awe-inspiring as some of the others I have seen when it comes to the athleticism and """never before seen""" type stuff. But the show was packed with a great story, amazing costumes, graphics, dancing, etc., and I loved every minute of it!
    \end{enumerate}
    } \\
    \hline
    {YelpP} & \parbox{\hsize}{Neuron 104:\\Unattractive store layout.} & \parbox{\hsize}{
    \begin{enumerate}
        \item I totally agree with Tina S. for such a large and beautiful store to be quite honest......the selection in a word.....SUCKS. The only reason I didn't give this store one star was because it is a very spacious store....but I think they waste a lot of space......and the customer service was excellent. However when you go into a Nike Store of any kind.....exception being the outlets.......there should be more than 7 or 8 NFL team Jerseys and T-Shirts in the place. I was extremely disappointed with that....and for that fact that is why I have never been a huge fan of Nike products or stores. Eat, Drink, and be Merry my Friends!!!!!
        \item This mall- eh It's not horrible, but it's a waste of time.  I visited from out of town and it was not worth my while. The stores were your typical """upscale""" shops, but good luck finding anything with the pacs of shoppers looking to score """deals""".  The only stores worth going to are Gap outlet and J Crew factory. I was excited when I saw H\&M but don't be fooled, it's not an outlet store so no """special""" deals there. Avoid the crowds, save the gas \$ and go elsewhere.  Pros: - I got 2 dresses at Gap outlet for less than \$20 Cons: - Crowded - Lack of selection - Not all stores are outlets even though this is an outlet mall - No food courts and when you put your credit card in the vending machine good luck getting your drink
        \item I made a few trips to this mall during our week in the Phoenix area.   The Nike Outlet was great, but otherwise, there weren't that many quality outlet stores.  Most (or so it seemed to me) of the stores in this mall are not outlets and there just weren't the deals that I was expecting.
        \item I hate to say it, but this mall is kind of ghetto. The layout is somewhat bizarre, and depending on which side you enter, you'd never know about the other side if you didn't look at a map and just decide to wander. The stores are really nothing special and if you seek high end stores, you're better off hitting the strip. What's really weird is the women's stores in there--they're either plus sizes, clubby looking stuff, or outright hooker uniforms. There is also a Macy's, JC Penny and Sears. Three stores I never buy anything from anyway. There is, however, a Cinnabon, and I LOVE Cinnabon....
        \item This mall is sad. You will actually feel bad for this mall. Only a couple shops are open and they are either shoe stores, clothing or cell phones. The food court doesn't make any sense and not very inviting. Also there wasn't a mrs. Fields cuz I was craving cookies. Lol Your better off going to the flea market for better stuff and cheaper prices!
    \end{enumerate}
    } \\
    \hline
    {AGnews} & \parbox{\hsize}{Neuron 20:\\sports events and achievements.} & \parbox{\hsize}{
    \begin{enumerate}
        \item Maddux Wins No. 302, Baker Wins No. 1,000 Greg Maddux pitched the Chicago Cubs into the lead in the NL wild-card race and gave Dusty Baker a win to remember. Maddux threw seven shutout innings for his 302nd career win, Baker got his 1,000th victory as a manager and Chicago beat the Montreal Expos 5-2 on Monday night...
        \item Colts Lead Pats Early in Third Quarter FOXBORO, Mass. - Peyton Manning reached the 25,000-yard passing mark faster than anyone but Dan Marino, and the Indianapolis Colts shredded the New England Patriots for a 17-13 halftime lead Thursday night...
        \item Davenport Advances at U.S. Open NEW YORK - Lindsay Davenport's summer of success stayed on course Thursday when the fifth-seeded former U.S. Open champion defeated Arantxa Parra Santonja 6-4, 6-2 and advanced to the third round of the season's final Grand Slam event...
        \item U.S. Men's Hoops Team Finally Gets a Rout ATHENS, Greece - The Americans got a taste of what it was like in the good ol' days. They finally played an opponent they were able to beat easily, routing Angola 89-53 Monday in their final preliminary game of the Olympic men's basketball tournament...
        \item U.S. Softball Team Wins, Closes in on Gold ATHENS, Greece - Right now, the Americans aren't just a Dream Team - they're more like the Perfect Team. Lisa Fernandez pitched a three-hitter Sunday and Crystl Bustos drove in two runs as the Americans rolled to their eighth shutout in eight days, 5-0 over Australia, putting them into the gold medal game...
    \end{enumerate}
    } \\
    \hline
    {AGnews} & \parbox{\hsize}{Neuron 16:\\human rights violations and advocacy.} & \parbox{\hsize}{
    \begin{enumerate}
        \item US soldier convicted of torture in Iraq A US military intelligence soldier in Iraq has been sentenced to 8 months in prison for taking part in torturing detainees in Abu Ghraib prison.
        \item Pinochet is ordered to stand trial for murder Augusto Pinochet, the former Chilean dictator, was ordered under house arrest yesterday, charged with kidnapping and murder dating back to his 17-year rule.
        \item Trial Date Set for Soldier at Abu Ghraib (AP) AP - A military judge ordered a U.S. Army reservist on Friday to stand trial Jan. 7 in Baghdad for allegedly abusing Iraq inmates at the Abu Ghraib prison outside Baghdad.
        \item Afghan court convicts US trio of torture KABUL, Afghanistan -- Three Americans -- led by a former Green Beret who boasted he had Pentagon support -- were found guilty yesterday of torturing Afghans in a private jail and were sentenced to prison.
        \item Soldier to Plead Guilty in Iraq Abuse Case (AP) AP - An Army reservist charged with abusing Iraqi prisoners plans to plead guilty at a court martial to four counts arising from the Abu Ghraib prison abuse scandal in a plea deal in which eight other counts will be dropped, his lawyer has said.
    \end{enumerate}
    } \\
    \hline
    {AGnews} & \parbox{\hsize}{Neuron 10:\\terrorism and security threats.} & \parbox{\hsize}{
    \begin{enumerate}
        \item Pakistan's top wanted terrorist killed Pakistani security forces Sunday killed the country's most wanted terrorist allegedly involved in an assassination attempt on President Pervez Musharrafand indicted in the murder of a US journalist.
        \item Al-Qaeda Group Kills a Second US Hostage in Iraq (Update3) An Iraqi group linked to al-Qaeda killed a second US hostage, Jack Hensley, and threatened to kill a British hostage unless Iraqi women detainees are freed, the group said on its Web site.
        \item Pakistan arrests key Al-Qaeda operative (AFP) AFP - Pakistani security forces have arrested a key Al-Qaeda operative wanted in connection with attacks on Christian targets and a failed bid to kill President Pervez Musharraf, an official said.
        \item Pakistan al-Qaeda suspect killed Pakistan says it has dealt a major blow to al-Qaeda's operations after its security forces shot dead the country's most wanted terror suspect.
        \item Seven suspected terrorists arrested in Spain Spain's Interior Minister says police have broken up a radical Muslim cell, plotting to bomb the country's National Court."
    \end{enumerate}
    } \\
    \hline
    {DBpedia} & \parbox{\hsize}{Neuron 174:\\words related to ship, car, train.} & \parbox{\hsize}{
    \begin{enumerate}
        \item USS England (DE-635) a Buckley-class destroyer escort of the United States Navy was named in honor of Ensign John C. England (1920–1941) who was killed in action aboard the battleship Oklahoma during the Japanese attack on Pearl Harbor on 7 December 1941.
        \item HMS Siren (most often referred to as Syren in contemporary records) was a sixth-rate post ship of the British Royal Navy in commission between 1745 and 1763 seeing action during the War of the Austrian Succession and the Seven Years' War.
        \item HMS Benbow was a Victorian era Admiral-class battleship of the British Royal Navy named for Admiral John Benbow.
        \item HMS Rackham was one of 93 ships of the Ham-class of inshore minesweepers. Their names were all chosen from villages ending in -ham. The minesweeper was named after Rackham in West Sussex.
        \item HMS Captain was a 74-gun third-rate ship of the line of the Royal Navy launched on 26 November 1787 at Limehouse. She served during the French revolutionary and Napoleonic Wars before being placed in harbour service in 1799. An accident caused her to burn and founder in 1813. Later that year she was raised and broken up.
    \end{enumerate}
    } \\
    \hline
    {DBpedia} & \parbox{\hsize}{Neuron 71:\\the artist's born date.} & \parbox{\hsize}{
    \begin{enumerate}
        \item Joanna Taylor (born 24 July 1978) is an English actress and former model.
        \item Jody Miller (born November 29 1941) is an American country music singer. Born as Myrna Joy Miller she was born in Phoenix Arizona and raised in Oklahoma.
        \item Priscilla Mitchell (born September 18 1941 in Marietta Georgia) was an American country music singer.
        \item Geoffrey Davies (born 15 December 1942 Leeds West Riding of Yorkshire) is a British actor.
        \item He was born in Asunción Paraguay on March 27 1950. Son of Carmen Emategui and Rodolfo Barreto.
    \end{enumerate}
    } \\
    \hline
    {DBpedia} & \parbox{\hsize}{Neuron 469:\\the publisher and imprint of the work.} & \parbox{\hsize}{
    \begin{enumerate}
        \item The Tameside Advertiser is a weekly newspaper which serves the Metropolitan Borough of Tameside Greater Manchester England. It is owned by Trinity Mirror plc. The paper has a sister paper The Glossop Advertiser which is also a freesheet but covers the bordering town of Glossop in Derbyshire. The main competitors to both papers are the Tameside Reporter and Glossop Chronicle which are both paid-for newspapers.
        \item Independent Tribune is a newspaper and based in Concord North Carolina covering Cabarrus County North Carolina. The newspaper is owned by Berkshire Hathaway. The Independent Tribune was formed with the merger of The Concord Tribune and The (Kannapolis) Daily Independent.It was originally a daily newspaper but changed to 3 days a week in 2009.
        \item The Livingston County Daily Press \& Argus is a daily newspaper published in Howell Michigan and owned by Gannett. 'As its name implies it covers news and sports within Livingston County and had offices in both Howell and Brighton. The Brighton office closed in December 2008. Its printing facility is located in Howell Township. It publishes every day except Saturday.
        \item The Anchorage Press is a free alternative weekly newspaper based in Anchorage Alaska and owned by Wick Communications.Established in 1992 by Bill Boulay Barry Bialik and Nick Coltman as the Anchorage Bypass it was renamed the Anchorage Press in 1994. It is published and distributed every Thursday with a circulation of approximately 25000. The paper was sold to Wick Communications Company in August 2006.
        \item The Imperial Valley Press (originally known as the Imperial Press) is a daily newspaper published in El Centro California. It has been owned by Schurz Communications of South Bend Indiana since 1965.The Imperial Valley Press features local news from all communities of the Imperial Valley and the Mexicali Baja California area as well as San Diego County and portions of southwestern Arizona. The newspaper focuses on local news sports and opinion pieces.
    \end{enumerate}
    } \\
\hline
\end{longtable}
\clearpage

\subsection{explanations from CB-LLM}
\label{sec:explanations}
In this section, we provide more explanations generated by our CB-LLM. We randomly select 3 samples and show the top 5 explanations for each dataset.
\LTcapwidth=0.98\textwidth
\begin{longtable}{|l|p{7.7cm}|p{6.7cm}|}
\caption {The explanations generated by CB-LLM w/ ACC for a given text sample. We show 3 random samples for each dataset.}
\vspace{-10pt}
\label{table:explanation} \\
\hline
    Dataset & Sample & Explanations \\
    \hline
    {SST2} & \parbox{\hsize}{Sample 260:\\a very witty take on change , risk and romance , and the film uses humour to make its points about ACC
    eptance and growth .} & \parbox{\hsize}{
    \begin{enumerate}
        \item Clever and unexpected humor.
        \item Charming characters.
        \item Stellar and diverse ensemble cast.
        \item Unique and well-developed characters.
        \item Captivating and layered character backstories.
    \end{enumerate}
    } \\
    \hline
        {SST2} & \parbox{\hsize}{Sample 1649:\\i was perplexed to watch it unfold with an astonishing lack of passion or uniqueness .} & \parbox{\hsize}{
    \begin{enumerate}
        \item Lack of tension-building scenes.
        \item Unexplained or unresolved mysteries.
        \item Uninspiring character deaths.
        \item Poorly executed voice-over narration.
        \item Lack of authentic cultural representation.
    \end{enumerate}
    } \\
    \hline
    {SST2} & \parbox{\hsize}{Sample 330:\\occasionally funny , always very colorful and enjoyably overblown in the traditional almodóvar style .} & \parbox{\hsize}{
    \begin{enumerate}
        \item Charming characters.
        \item Clever and unexpected humor.
        \item Stunning and exotic locations.
        \item Stellar and diverse ensemble cast.
        \item Unique and well-developed characters.
    \end{enumerate}
    } \\
    \hline
    {YelpP} & \parbox{\hsize}{Sample 21864:\\These guys are money grubbing.  What WAS a \$25 haircut just jumped up to a \$32 haircut.  It's just a haircut for God's sake!  I'm going elsewhere.} & \parbox{\hsize}{
    \begin{enumerate}
        \item Poor customer service.
        \item Unattractive store layout.
        \item Rude staff.
        \item Hidden fees.
        \item Overpriced.
    \end{enumerate}
    } \\
    \hline
    {YelpP} & \parbox{\hsize}{\vspace{0.1cm}Sample 34857:\\This place has something for everyone.  My wife and I started going there out of convenience before attending a movie at the South Pointe.  But then we continued going back because we liked the food and the staff is very helpful.  This most recent visit I had sushi for the first time and it was very good - and reasonably priced.  We have company coming and are going to make it one of our stops on their visit.\vspace{0.05cm}} & \parbox{\hsize}{
    \begin{enumerate}
        \item Welcoming and friendly staff.
        \item Clean and inviting ambiance.
        \item Amazing flavors.
        \item Great warranty and support.
        \item Delicious food.
    \end{enumerate}
    } \\
    \hline
    {YelpP} & \parbox{\hsize}{\vspace{0.1cm}Sample 10736:\\One of the few Cirque du Soleil that follow a story line, so if you are looking for a Cirque du Soleil show and a story this is the one to see. Although it strays a bit from the traditional style of Cirque du Soleil, it is still sure to please. We were fortunate enough to be able to purchase front section tickets for 50\% off AMAZING deal! (End of summer special). KA is the show which it is the stage that is at the center of attention. It uses a sectional stage that is fully mobile it rotates and moves on a 3D axis it really adds another level of excitement to the show. I would not recommend this as anyone's first Cirque du Soleil show but for a any repeat or veteran Cirque du Soleil viewer this must make it onto your \""Seen it\"" list.\vspace{0.05cm}} & \parbox{\hsize}{
    \begin{enumerate}
        \item Engaging performances.
        \item Clean and inviting ambiance.
        \item Interactive experiences.
        \item Engaging podcasts.
        \item Welcoming and friendly staff.
    \end{enumerate}
    } \\
    \hline
    {AGnews} & \parbox{\hsize}{Sample 3058:\\Mobile phone network reaches last of China's ethnic minorities (AFP) AFP - China has brought its mobile phone network to the last of its ethnic minority regions previously cut off from communication with the outside world, state media reported.} & \parbox{\hsize}{
    \begin{enumerate}
        \item telecommunications and 5G technology.
        \item tech giants and major industry players.
        \item consumer electronics and gadgets.
        \item emerging technologies and startups.
        \item words related to technical devices.
    \end{enumerate}
    } \\
    \hline
    {AGnews} & \parbox{\hsize}{Sample 6124:\\Van Gogh's murder brings out Holland's contradictions The murder of Dutch filmmaker Theo van Gogh by a young Muslim of Moroccan descent has shaken Holland to its very foundations. To most people, including the Dutch, the killing and its violent} & \parbox{\hsize}{
    \begin{enumerate}
        \item human rights violations and advocacy.
        \item terrorism and security threats.
        \item words related to war, conflict.
        \item international aid and humanitarian efforts.
        \item public health crises and pandemics.
    \end{enumerate}
    } \\
    \hline
    {AGnews} & \parbox{\hsize}{Sample 1035:\\Orioles 8, Devil Rays 0 Javy Lopez drove in four runs, Daniel Cabrera became the first rookie to win 10 games this season, and the Baltimore Orioles held the Tampa Bay Devil Rays to two hits in an 8-0 victory.} & \parbox{\hsize}{
    \begin{enumerate}
        \item team rankings and standings.
        \item fan reactions and opinions.
        \item record-breaking performances.
        \item athlete comebacks after injury.
        \item name of sports stars.
    \end{enumerate}
    } \\
    \hline
    {DBpedia} & \parbox{\hsize}{\vspace{0.1cm}Sample 52170:\\Narthecium is a genus of flowering plants. This genus was traditionally treated as belonging to the family Liliaceae but the APG II system of 2003 placed it in the family Nartheciaceae.The global distribution of the genus is widely disjunct - 1 species in Asia 1-5 species in Europe (see Narthecium ossifragum and 2 species in North America. Narthecium americanum is a candidate for listing under the federal Endangered Species Act in the United States.\vspace{0.1cm}} & \parbox{\hsize}{
    \begin{enumerate}
        \item The botanical classification of the plant.
        \item the name of the plant.
        \item The native habitat of the plant.
        \item the genus or family of plant.
        \item The plant's contribution to biodiversity.
    \end{enumerate}
    } \\
    \hline
    {DBpedia} & \parbox{\hsize}{Sample 32678:\\Pemberton's Headquarters also known as Willis-Cowan House is a two-story brick house that served as the headquarters for Confederate General John C. Pemberton during most of the 47 day siege of Vicksburg and the site where he decided to surrender the city to Union General Ulysses S. Grant on July 4 1863.During the 1960s the building housed a kindergarten associated with Vicksburg Catholic School (St.} & \parbox{\hsize}{
    \begin{enumerate}
        \item the location of the building.
        \item The historical significance of the building.
        \item the name of the building.
        \item the built date of the building.
        \item The cultural or artistic significance of the building.
    \end{enumerate}
    } \\
    \hline
    {DBpedia} & \parbox{\hsize}{\vspace{0.1cm}Sample 12750:\\Disma Fumagalli (born Inzago September 8 1826 - died Milan March 9 1893) was an Italian composer and teacher of music. He was a graduate of the Milan Conservatory where he began teaching piano in 1853. He composedmore than 300 études for piano as well as other exercises; he also wrote a concerto for piano and string orchestra. Fumagalli's brothers Carlo Polibio Adolfo and Luca were all composers.\vspace{0.05cm}} & \parbox{\hsize}{
    \begin{enumerate}
        \item the artist's born date
        \item The artist's cultural significance.
        \item The artist's famous collaborations.
        \item The artist's notable achievements.
        \item The artist's early influences.
    \end{enumerate}
    } \\
\hline
\end{longtable}
\clearpage

\subsection{Details of prompting ChatGPT}
\label{sec:detailed prompts}
In this section, We provide the details of how we prompt ChatGPT to acquire the concept set. We use four human-designed concepts as examples for in-context learning. This prompting style requires only $n$ queries to ChatGPT to obtain the full concept set and can be done efficiently through the web interface provided by OpenAI. The full prompts are shown in \ref{table:prompt}.
\begin{longtable}{|l|p{1.8cm}|p{12.7cm}|}
\caption {The designed prompts for each dataset and class.}
\vspace{-10pt}
\label{table:prompt} \\
\hline
    Dataset & Class & Prompt \\
    \hline
    {SST2} & negative & \parbox{\hsize}{\vspace{0.3cm}Here are some examples of key features that are often present in a negative movie rating. Each feature is shown between the tag <example></example>. \\
    <example>Flat or one-dimensional characters.</example> \\
    <example>Uninteresting cinematography.</example> \\
    <example>Lack of tension-building scenes.</example> \\
    <example>Lack of emotional impact.</example> \\
    List 100 other different important features that are often present in a negative movie rating. Need to follow the template above, i.e. <example>features</example>.\vspace{0.3cm}} \\
    \hline
    {SST2} & positive & \parbox{\hsize}{\vspace{0.3cm}Here are some examples of key features that are often present in a positive movie rating. Each feature is shown between the tag <example></example>. \\
    <example>Engaging plot.</example> \\
    <example>Strong character development.</example> \\
    <example>Great humor.</example> \\
    <example>Clever narrative structure.</example> \\
    List 100 other different important features that are often present in a positive movie rating. Need to follow the template above, i.e. <example>features</example>.\vspace{0.3cm}} \\
    \hline
    {YelpP} & negative & \parbox{\hsize}{\vspace{0.3cm}Here are some examples of key features that are often present in a negative Yelp review with lower star ratings (e.g., 1 or 2 stars). Each feature is shown between the tag <example></example>. \\
    <example>Overpriced.</example> \\
    <example>Unappetizing food.</example> \\
    <example>Unprofessional service.</example> \\
    <example>broken products.</example> \\
    The reviews fall into the following categories: Food, Automotive, Home Services, Entertainment, Medical, Hotels, Financial Services, Media, Parking, Clothing, Electronic devices, and Cleaning. List 100 other different important features that are often present in a negative Yelp review with lower star ratings (e.g., 1 or 2 stars). Need to follow the template above, i.e. <example>features</example>.\vspace{0.3cm}} \\
    \hline
    {YelpP} & positive & \parbox{\hsize}{\vspace{0.3cm}Here are some examples of key features that are often present in a positive Yelp review with higher star ratings (e.g., 4 or 5 stars). Each feature is shown between the tag <example></example>. \\
    <example>Delicious food.</example> \\
    <example>Outstanding service.</example> \\
    <example>Great value for the price.</example> \\
    <example>high quality products.</example> \\
    The reviews fall into the following categories: Food, Automotive, Home Services, Entertainment, Medical, Hotels, Financial Services, Media, Parking, Clothing, Electronic devices, and Cleaning. List 100 other different important features that are often present in a positive Yelp review with higher star ratings (e.g., 4 or 5 stars). Need to follow the template above, i.e. <example>features</example>.\vspace{0.3cm}} \\
    \hline
    {AGnews} & world & \parbox{\hsize}{\vspace{0.3cm}Here are some examples of key features that are often present in worldwide news. Each feature is shown between the tag <example></example>.\\
    <example>words related to country and place.</example> \\
    <example>political stunts taken by governments.</example> \\
    <example>global issues.</example> \\
    <example>words related to war, conflict.</example> \\
    List 50 other important features that are often present in worldwide news. Need to follow the template above, i.e. <example>features</example>.\vspace{0.3cm}}\\
    \hline
    {AGnews} & sports & \parbox{\hsize}{\vspace{0.3cm}Here are some examples of key features that are often present in sport news. Each feature is shown between the tag <example></example>. \\
    <example>name of sports stars.</example> \\
    <example>words related to game, competition.</example> \\
    <example>ball games like baseball, basketball.</example> \\
    <example>name of sport teams.</example> \\
    List 50 other important features that are often present in sport news. Need to follow the template above, i.e. <example>features</example>.\vspace{0.3cm}} \\
    \hline
    {AGnews} & business & \parbox{\hsize}{\vspace{0.3cm}Here are some examples of key features that are often present in business and financial news. Each feature is shown between the tag <example></example>. \\
    <example>words related to currency, money.</example> \\
    <example>the numerical amount of dollars.</example> \\
    <example>the symbol like \$.</example> \\
    <example>words related to stock, Portfolio.</example> \\
    List 50 other important features that are often present in business and financial news. Need to follow the template above, i.e. <example>features</example>.\vspace{0.3cm}} \\
    \hline
    {AGnews} & \parbox{\hsize}{science/\\technology} & \parbox{\hsize}{\vspace{0.3cm}Here are some examples of key features that are often present in news related to science and technology. Each feature is shown between the tag <example></example>.\\
    <example>name of scientists or the word scientists.</example> \\
    <example>words related to technical devices.</example> \\
    <example>words related to universe, space, planet.</example> \\
    <example>words related to the natural landscape.</example> \\
    List 50 other important features that are often present in news related to science and technology. Need to follow the template above, i.e. <example>features</example>.\vspace{0.3cm}} \\
    \hline
    {DBpedia} & company & \parbox{\hsize}{\vspace{0.3cm}Here are some examples of key features that are often present when introducing a company. Each feature is shown between the tag <example></example>. \\
    <example>the name of the company.</example> \\
    <example>the location of the company</example> \\
    <example>the founding year of the company</example> \\
    <example>words related to organization, group.</example> \\
    List 30 other important features that are often present when introducing a company. Need to follow the template above, i.e. <example>features</example>.\vspace{0.3cm}}\\
    \hline
    {DBpedia} & \parbox{\hsize}{educational\\institution} & \parbox{\hsize}{\vspace{0.3cm}Here are some examples of key features that are often present when introducing an educational institution. Each feature is shown between the tag <example></example>. \\
    <example>the name of the school.</example> \\
    <example>the location of the school</example> \\
    <example>the founding year of the school</example> \\
    <example>words related to college, university.</example> \\
    List 30 other important features that are often present when introducing an educational institution. Need to follow the template above, i.e. <example>features</example>.\vspace{0.3cm}}\\
    \hline
    {DBpedia} & artist & \parbox{\hsize}{\vspace{0.3cm}Here are some examples of key features that are often present when introducing an artist. Each feature is shown between the tag <example></example>. \\
    <example>the artist's name.</example> \\
    <example>the artist's works</example> \\
    <example>the artist's born date</example> \\
    <example>words related to music, painting.</example> \\
    List 30 other important features that are often present when introducing an artist. Need to follow the template above, i.e. <example>features</example>.\vspace{0.3cm}}\\
    \hline
    {DBpedia} & athlete & \parbox{\hsize}{\vspace{0.3cm}Here are some examples of key features that are often present when introducing an athlete or sports star. Each feature is shown between the tag <example></example>. \\
    <example>the athlete's or sports stars' name.</example> \\
    <example>the sport the athlete plays (e.g. football, basketball).</example> \\
    <example>the athlete's or sports stars' born date</example> \\
    <example>words related to ball games, competition.</example> \\
    List 30 other important features that are often present when introducing an athlete or sports star. Need to follow the template above, i.e. <example>features</example>.\vspace{0.3cm}}\\
    \hline
    {DBpedia} & office holder & \parbox{\hsize}{\vspace{0.3cm}Here are some examples of key features that are often present when introducing an office holder. Each feature is shown between the tag <example></example>. \\
    <example>the office holder's name.</example> \\
    <example>the office holder's position.</example> \\
    <example>the office holder's born date</example> \\
    <example>words related to politician, businessman.</example> \\
    List 30 other important features that are often present when introducing an office holder. Need to follow the template above, i.e. <example>features</example>.\vspace{0.3cm}}\\
    \hline
    {DBpedia} & transportation & \parbox{\hsize}{\vspace{0.3cm}Here are some examples of key features that are often present when introducing transportation. Each feature is shown between the tag <example></example>. \\
    <example>the model type of the transportation or vehicle.</example> \\
    <example>the production date of the transportation or vehicle.</example> \\
    <example>the functions of the transportation or vehicle.</example> \\
    <example>words related to ship, car, train.</example> \\
    List 30 other important features that are often present when introducing transportation. Need to follow the template above, i.e. <example>features</example>.\vspace{0.3cm}}\\
    \hline
    {DBpedia} & building & \parbox{\hsize}{\vspace{0.3cm}Here are some examples of key features that are often present when introducing a building. Each feature is shown between the tag <example></example>. \\
    <example>the name of the building.</example> \\
    <example>the built date of the building.</example> \\
    <example>the location of the building.</example> \\
    <example>words related to the type of the building (e.g. church, historic house, park, resort).</example> \\
    List 30 other important features that are often present when introducing a building. Need to follow the template above, i.e. <example>features</example>.\vspace{0.3cm}}\\
    \hline
    {DBpedia} & natural place & \parbox{\hsize}{\vspace{0.3cm}Here are some examples of key features that are often present when introducing a natural place. Each feature is shown between the tag <example></example>. \\
    <example>the name of the natural place.</example> \\
    <example>the length or height of the natural place.</example> \\
    <example>the location of the natural place.</example> \\
    <example>words related to mountain, river.</example> \\
    List 30 other important features that are often present when introducing a natural place. Need to follow the template above, i.e. <example>features</example>.\vspace{0.3cm}}\\
    \hline
    {DBpedia} & village & \parbox{\hsize}{\vspace{0.3cm}Here are some examples of key features that are often present when introducing a village. Each feature is shown between the tag <example></example>. \\
    <example>the name of the village.</example> \\
    <example>the population of the village.</example> \\
    <example>the census of the village.</example> \\
    <example>words related to district, families.</example> \\
    List 30 other important features that are often present when introducing a village. Need to follow the template above, i.e. <example>features</example>.\vspace{0.3cm}}\\
    \hline
    {DBpedia} & animal & \parbox{\hsize}{\vspace{0.3cm}Here are some examples of key features that are often present when introducing a kind of animal. Each feature is shown between the tag <example></example>. \\
    <example>the species of the animal.</example> \\
    <example>the habitat of the animal.</example> \\
    <example>the type of the animal (e.g. bird, insect, moth).</example> \\
    <example>words related to genus, family.</example> \\
    List 30 other important features that are often present when introducing a kind of animal. Need to follow the template above, i.e. <example>features</example>.\vspace{0.3cm}}\\
    \hline
    {DBpedia} & plant & \parbox{\hsize}{\vspace{0.3cm}Here are some examples of key features that are often present when introducing a kind of plant. Each feature is shown between the tag <example></example>. \\
    <example>the name of the plant.</example> \\
    <example>the genus or family of plant.</example> \\
    <example>the place where the plant was found.</example> \\
    <example>words related to grass, herb, flower.</example> \\
    List 30 other important features that are often present when introducing a kind of plant. Need to follow the template above, i.e. <example>features</example>.\vspace{0.3cm}}\\
    \hline
    {DBpedia} & album & \parbox{\hsize}{\vspace{0.3cm}Here are some examples of key features that are often present when introducing an album. Each feature is shown between the tag <example></example>. \\
    <example>the name of the album.</example> \\
    <example>the type of music, instrument.</example> \\
    <example>the release date of the album.</example> \\
    <example>words related to band, studio.</example> \\
    List 30 other important features that are often present when introducing an album. Need to follow the template above, i.e. <example>features</example>.\vspace{0.3cm}}\\
    \hline
    {DBpedia} & film & \parbox{\hsize}{\vspace{0.3cm}Here are some examples of key features that are often present when introducing a film. Each feature is shown between the tag <example></example>. \\
    <example>the name of the film.</example> \\
    <example>the maker or producer of the film.</example> \\
    <example>the type of the film (e.g. drama, science fiction, comedy, cartoon, animation).</example> \\
    <example>words related to TV, video.</example> \\
    List 30 other important features that are often present when introducing a film. Need to follow the template above, i.e. <example>features</example>.\vspace{0.3cm}}\\
    \hline
    {DBpedia} & \parbox{\hsize}{written\\work} & \parbox{\hsize}{\vspace{0.3cm}Here are some examples of key features that are often present when introducing a written work. Each feature is shown between the tag <example></example>. \\
    <example>the name of the written work.</example> \\
    <example>the author of the film.</example> \\
    <example>the type of the written work (e.g. novel, manga, journal).</example> \\
    <example>words related to book.</example> \\
    List 30 other important features that are often present when introducing a written work. Need to follow the template above, i.e. <example>features</example>.\vspace{0.3cm}}\\
\hline
\end{longtable}
\clearpage

%%%%%%%%%%%%%%%%%%%%%%%%%%%%%%%%%%%%%%%%%%%%%%%%%%%%%%%%%%%%%%%%%%%%%%%%%%%%%%%
%%%%%%%%%%%%%%%%%%%%%%%%%%%%%%%%%%%%%%%%%%%%%%%%%%%%%%%%%%%%%%%%%%%%%%%%%%%%%%%

\end{document}